# An Analysis of the t-SNE Algorithm for Data Visualization[*]

Sanjeev Arora
Princeton University
arora@cs.princeton.edu

Wei Hu
Princeton University
huwei@cs.princeton.edu

Pravesh K. Kothari
Princeton University & Institute for Advanced Study
kothari@cs.princeton.edu

## Abstract

A first line of attack in exploratory data analysis is *data visualization*, i.e., generating a 2-dimensional representation of data that makes *clusters* of similar points visually identifiable. Standard Johnson-Lindenstrauss dimensionality reduction does not produce data visualizations. The *t-SNE* heuristic of van der Maaten and Hinton, which is based on non-convex optimization, has become the *de facto* standard for visualization in a wide range of applications.

This work gives a formal framework for the problem of data visualization – finding a 2-dimensional embedding of clusterable data that correctly separates individual clusters to make them visually identifiable. We then give a rigorous analysis of the performance of t-SNE under a natural, deterministic condition on the "ground-truth" clusters (similar to conditions assumed in earlier analyses of clustering) in the underlying data. These are the first provable guarantees on t-SNE for constructing good data visualizations.

We show that our deterministic condition is satisfied by considerably general probabilistic generative models for clusterable data such as mixtures of well-separated log-concave distributions. Finally, we give theoretical evidence that t-SNE provably succeeds in *partially* recovering cluster structure even when the above deterministic condition is not met.

## 1 Introduction

Many scientific applications, especially those involving exploratory data analysis, rely on visually identifying high-level qualitative structures in the data, such as clusters or groups of similar points. This is not easy since the data of interest is usually high-dimensional and it is unclear how to capture the qualitative cluster structure in a 2-dimensional visualization. For example, linear dimensionality reduction techniques (e.g., data oblivious Johnson-Lindenstrauss (JL) embedding or data-dependent embedding using PCA) are incapable of reducing dimension down to 2 in any meaningful way (see Figure 1) - they merge distinct clusters into a uniform-looking sea of points.

In 2008, van der Maaten and Hinton (2008) introduced a nonlinear algorithm, *t-Distributed Stochastic Neighbor Embedding or t-SNE* (an improvement over the earlier SNE algorithm of Hinton and Roweis (2002)) for this task, which has become the de facto standard (see Figure 1c) for visualizing high-dimensional datasets with diverse applications such as computer security (Gashi et al., 2009), music analysis (Hamel and Eck, 2010), cancer biology (Abdelmoula et al., 2016) and bioinformatics (Wallach and Lilien, 2009).

---

[*]Accepted for presentation at Conference on Learning Theory (COLT) 2018.

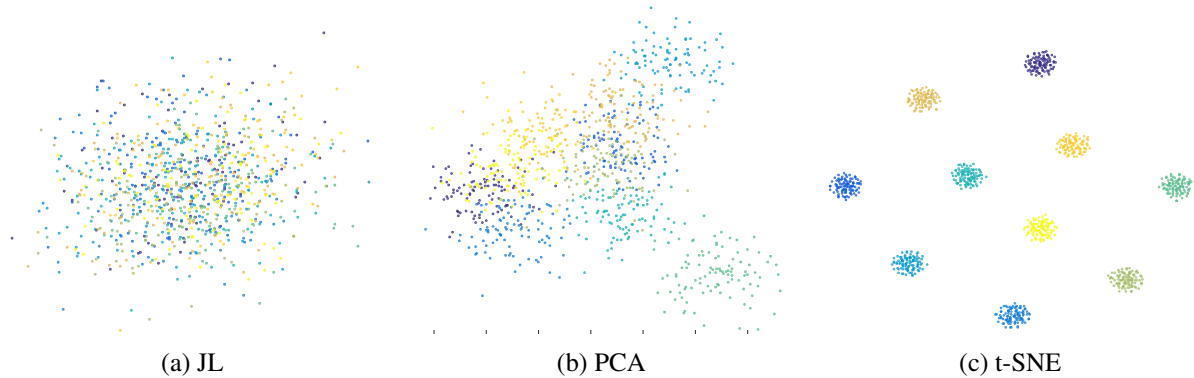

Figure 1: 2D embeddings of a mixture of 10 Gaussians with pairwise center separation $0.5\times$radius via: (a) random projection (JL), (b) projection to the subspace of top 2 singular vectors (PCA), (c) t-SNE.

At a high level, t-SNE (like SNE) chooses two similarity measures between pairs of points - one for the high dimensional data and one for the 2-dimensional embedding. It then attempts to construct a 2-dimensional embedding that minimizes the KL divergence between the vector of similarities between pairs of points in the original dataset and the similarities between pairs of points in the embedding. This is a non-convex optimization problem and t-SNE employs gradient descent with random initialization (along with other tricks such as *early exaggeration*) to compute a reasonable solution to it. See Section 2 for details.

Of course, non-convex optimization drives much of today's progress in machine learning and data science, and thus poses a rich set of theoretical questions. Researchers have managed to rigorously analyze non-convex optimization algorithms in a host of settings (Dasgupta, 1999; Arora et al., 2012, 2014; Bhojanapalli et al., 2016; Ge et al., 2015a; Sun et al., 2017; Ge et al., 2017, 2016; Park et al., 2017). These analyses usually involve making clean assumptions about the structure of data, usually with a generative model. The goal of the current paper is to rigorously analyze t-SNE in a similar vein.

At the outset such a project runs into definitional issues about what a good *visualization* of clustering is. Many such issues are inherited from well-known issues in formalizing the goals of clustering (Kleinberg, 2002). In theoretical studies of clustering, such issues were sidestepped by going with a standard clustering formalization and assuming that data come with an (unknown) ground-truth clustering (for instance, mixtures of Gaussians, $k$-means, etc.). We make similar assumptions and assume that our goal is to produce a 2-dimensional embedding such that the points in the same clusters are noticeably closer together compared with points in different clusters. Under some of these standard models we show that t-SNE provably succeeds in computing a good visualization.

We emphasize that the focus of this paper is on formalizing the notion of visualization and providing a theoretical analysis of t-SNE. We do not advocate for t-SNE over other visualization methods.

We now begin by describing our formalization of the visualization problem followed by describing our results that give the first provable guarantees on t-SNE for computing visualization of clusterable data.

**Formalizing Visualization.** We assume that we are given a collection of points $\mathcal{X} = \{x_1, x_2, \ldots, x_n\} \subset \mathbb{R}^d$ and that there exists a "ground-truth" clustering described by a partition $\mathcal{C}_1, \mathcal{C}_2, \ldots, \mathcal{C}_k$ of $[n]$ into $k$ clusters.

A visualization is described by a 2-dimensional embedding $\mathcal{Y} = \{y_1, y_2, \ldots, y_n\} \subseteq \mathbb{R}^2$ of $\mathcal{X}$, where each $x_i \in \mathcal{X}$ is mapped to the corresponding $y_i \in \mathcal{Y}$. Intuitively, a cluster $\mathcal{C}_\ell$ in the original data is *visualized* if the corresponding points in the 2-dimensional embedding $\mathcal{Y}$ are well-separated from all the rest. The following definition formalizes this idea.

**Definition 1.1** (Visible cluster). *Let $\mathcal{Y}$ be a 2-dimensional embedding of a dataset $\mathcal{X}$ with ground-truth*

*clustering* $\mathcal{C}_1, \ldots, \mathcal{C}_k$. *Given* $\epsilon \geq 0$, *a cluster* $\mathcal{C}_\ell$ *in* $\mathcal{X}$ *is said to be* $(1-\epsilon)$-*visible in* $\mathcal{Y}$ *if there exist* $\mathcal{P}, \mathcal{P}_{\text{err}} \subseteq [n]$ *such that:*

1. $|(\mathcal{P} \setminus \mathcal{C}_\ell) \cup (\mathcal{C}_\ell \setminus \mathcal{P})| \leq \epsilon \cdot |\mathcal{C}_\ell|$, $|\mathcal{P}_{\text{err}}| \leq \epsilon n$, *and*

2. *for every* $i, i' \in \mathcal{P}$ *and* $j \in [n] \setminus (\mathcal{P} \cup \mathcal{P}_{\text{err}})$, $\|y_i - y_{i'}\| \leq \frac{1}{2}\|y_i - y_j\|$.

*In such a case, we say that* $\mathcal{P}$ $(1-\epsilon)$-*visualizes* $\mathcal{C}_i$ *in* $\mathcal{Y}$.

It is now easy to define when $\mathcal{Y}$ is a good *visualization* - we ask that every cluster $\mathcal{C}_\ell$ in the dataset $\mathcal{X}$ is visualized in $\mathcal{Y}$.

**Definition 1.2** (Visualization). *Let* $\mathcal{Y}$ *be a 2-dimensional embedding of a dataset* $\mathcal{X}$ *with ground-truth clustering* $\mathcal{C}_1, \ldots, \mathcal{C}_k$. *Given* $\epsilon \geq 0$, *we say that* $\mathcal{Y}$ *is a* $(1-\epsilon)$-*visualization of* $\mathcal{X}$ *if there exists a partition* $\mathcal{P}_1, \mathcal{P}_2, \ldots, \mathcal{P}_k, \mathcal{P}_{\text{err}}$ *of* $[n]$ *such that:*

(i) *For each* $i \in [k]$, $\mathcal{P}_i$ $(1-\epsilon)$-*visualizes* $\mathcal{C}_i$ *in* $\mathcal{Y}$, *and*

(ii) $|\mathcal{P}_{\text{err}}| \leq \epsilon n$.

*In particular, when* $\epsilon = 0$, *we say that* $\mathcal{Y}$ *is a* full visualization *of* $\mathcal{X}$.

**Remark.** *Note that this formalization of visualization should be considered a first cut, since ultimately human psychology must come into play. For instance, humans may reasonably visualize two parallel lines as two clusters, but these violate our definition.*

*A natural question is whether clustering inferred from a visualization is unique. Our definition above does not guarantee this. Indeed, this is inherently impossible and relates to the ambiguity in the definition of clustering: for example, it can be impossible to determine whether a given set of points should be viewed as one cluster or two different smaller clusters. See Figure 2 for an example.*

*It is, however, not hard to establish that under an additional assumption that the size (number of points) of any cluster is smaller than twice the size of any other, full visualization as defined in Definition 1.2 uniquely determines a clustering.*

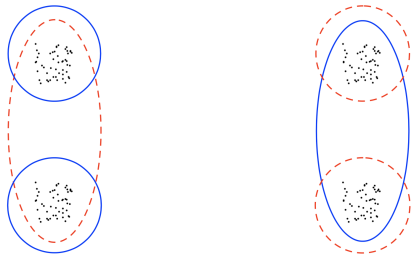

Figure 2: If we knew that there are 3 clusters in the original data, the blue and red outlines denote equally valid guesses for the underlying clustering based on the above visualization.

In order to study fine-grained behaviors of t-SNE, we also define a weaker variant of visualization where at least one cluster is visualized.

**Definition 1.3** (Partial visualization). *Given* $\epsilon \geq 0$, *we say that* $\mathcal{Y}$ *is a* $(1-\epsilon)$-*partial visualization of* $\mathcal{X}$ *if there exists a subset* $\mathcal{P} \subseteq [n]$ *such that* $\mathcal{P}$ $(1-\epsilon)$-*visualizes* $\mathcal{C}_\ell$ *for some* $\ell \in [k]$.

## 1.1 Our Results

Our main result identifies a simple deterministic condition on the clusterable data under which t-SNE provably succeeds in computing a full visualization.

**Definition 1.4** (Well-separated, spherical data). *Let $\mathcal{X} = \{x_1, x_2, \ldots, x_n\} \subset \mathbb{R}^d$ be clusterable data with $\mathcal{C}_1, \mathcal{C}_2, \ldots, \mathcal{C}_k$ defining the individual clusters such that for each $\ell \in [k]$, $|\mathcal{C}_\ell| \geq 0.1(n/k)$. We say that $\mathcal{X}$ is $\gamma$-spherical and $\gamma$-well-separated if for some $b_1, b_2, \ldots, b_k > 0$, we have:*

1. *$\gamma$-**Spherical:*** *For any $\ell \in [k]$ and $i, j \in \mathcal{C}_\ell$ ($i \neq j$), we have $\|x_i - x_j\|^2 \geq \frac{b_\ell}{1+\gamma}$, and for any $i \in \mathcal{C}_\ell$ we have $\left|\left\{j \in \mathcal{C}_\ell \setminus \{i\} : \|x_i - x_j\|^2 \leq b_\ell\right\}\right| \geq 0.51|C_\ell|$.*

2. *$\gamma$-**Well-separated:*** *For any $\ell, \ell' \in [k]$ ($\ell \neq \ell'$), $i \in \mathcal{C}_\ell$ and $j \in \mathcal{C}_{\ell'}$ we have $\|x_i - x_j\|^2 \geq (1 + \gamma \log n) \max\{b_\ell, b_{\ell'}\}$.*

The first condition asks for the distances between points in the same cluster ("intra-cluster distances") to be concentrated around a single value (with $\gamma$ controlling the "amount" of concentration). The second condition requires that the distances between two points from different clusters should be somewhat larger than the intra-cluster distances for each of the two clusters involved. In addition, we require that none of the clusters has too few points. Such assumptions are satisfied by well-studied probabilistic generative models for clusterable data such as mixture of Gaussians and more generally, mixture of log-concave distributions, and have been used in previous work (Dasgupta, 1999; Arora and Kannan, 2005) studying "distance-based" clustering algorithms.

For spherical and well-separated data, our main theorem below shows that t-SNE with early exaggeration succeeds in finding a full visualization.

**Theorem 1.5** (Informal, see Theorem 3.1 for a formal version). *Let $\mathcal{X} = \{x_1, x_2, \ldots, x_n\} \subset \mathbb{R}^d$ be $\gamma$-spherical and $\gamma$-well-separated clusterable data with $\mathcal{C}_1, \mathcal{C}_2, \ldots, \mathcal{C}_k$ defining the individual clusters. Then, t-SNE with early exaggeration on input $\mathcal{X}$ outputs a full visualization of $\mathcal{X}$ with high probability.*

**Proof Technique.** At a high level, t-SNE starts with a randomly initialized embedding and makes iterative gradient updates to it. The analysis thus demands understanding the effect of this update rule to the embedding of the high-dimensional points as a function of whether they lie in the same cluster or not. In a recent work, Linderman and Steinerberger (2017) established a "shrinkage" result for this update rule - they showed that points in the same cluster move towards each other under some mild conditions, that is, the embedding of any cluster "shrinks" as the iterations proceed. This result, however, is insufficient to establish that t-SNE succeeds in finding a full visualization as it does not rule out multiple clusters merging into each other.

We resort to a more fine-grained analysis built on the one by Linderman and Steinerberger (2017) and obtain an update rule for the *centroids* of the embeddings of all underlying clusters. This allows us to track the changes to the positions of the centroids and show that the distance between distinct centroids remains *lower-bounded* whenever the data is $\gamma$-spherical and $\gamma$-well-separated. Combined with the shrinkage result for points in the same cluster, this implies that t-SNE outputs a full visualization of the data.

Our analysis implicitly relies on the update rule in t-SNE closely mimicking those appearing in the well-studied *noisy power method* (with non-random noise). In Section 3.3, we make this connection explicit and show that the behavior of t-SNE (with early exaggeration) on $\gamma$-spherical and well-separated data can in fact be closely approximated by power method run on a natural matrix of pairwise similarities.

**Application to Visualizing Mixture Models.** Mixture of Gaussians and more generally, mixture of log-concave distributions, are well-studied probabilistic generative models for clusterable data. As an immediate application of our main theorem above, we show that t-SNE produces a full visualization for data generated according to such models. Before describing the result, we quickly recall the definition of mixture of log-concave distributions.

A distribution $\mathcal{D}$ with density function $f$ on $\mathbb{R}^d$ is said to be *log-concave* if $\log(f)$ is a concave function. $\mathcal{D}$ is said to be *isotropic* if its covariance is $I$. Many natural distributions including Gaussian distributions and the uniform distribution on any convex set are log-concave.

A mixture of $k$ log-concave distributions is described by $k$ positive mixing weights $w_1, w_2, \ldots, w_k$ ($\sum_{\ell=1}^{k} w_\ell = 1$) and $k$ log-concave distributions $\mathcal{D}_1, \ldots \mathcal{D}_k$ in $\mathbb{R}^d$. To sample a point from this model, we pick cluster $\ell$ with probability $w_\ell$ and draw $x$ from $\mathcal{D}_\ell$.

Theorem 1.5 immediately implies that t-SNE constructs a full visualization for data generated from a mixture of isotropic Gaussians or log-concave distributions with well-separated means. For isotropic Gaussians, the required pairwise separation between means is $\tilde{\Omega}(d^{1/4})$. For more general isotropic log-concave distributions, we require that the means be separated by $\tilde{\Omega}(d^{5/12})$.

Observe that the radius of samples from an isotropic log-concave distribution is $\approx d^{1/2}$ - thus, t-SNE succeeds in constructing 2D visualizations for clustering models far below the separation at which the clusters are non-overlapping. This is in stark contrast to standard linear dimensionality reduction techniques such as the Johnson-Lindenstrauss embedding that require mean separation of $\Omega(d^{1/2})$ to construct 2D visualizations that correctly separate 99% of points.

**Corollary 1.6** (Informal, see Corollary 3.11 for details). *Let $\mathcal{X} = \{x_1, \ldots, x_n\} \subset \mathbb{R}^d$ be i.i.d. samples from an equal-weighted mixture of $k$ isotropic Gaussians in $\mathbb{R}^d$ with every pair of distinct means separated by $\tilde{\Omega}(d^{1/4})$. Then, with high probability t-SNE with early exaggeration on input $\mathcal{X}$ outputs a full visualization of $\mathcal{X}$. Moreover, the same result holds for mixture of isotropic log-concave distributions with every pair of distinct means separated by $\tilde{\Omega}(d^{5/12})$.*

**Remark.** *Our result actually holds for a larger subclass of mixtures of* non-isotropic *log-concave distributions that may not be equal-weighted. See Corollary 3.11 for details. Mixture of log-concave distributions is among the weakest assumptions under which clustering algorithms with provable guarantees have been designed (Arora and Kannan, 2005; Vempala and Wang, 2004). We show that the t-SNE heuristic can visualize clusters under assumptions similar to the more sophisticated methods in previous theoretical work.*

Finally, we show that even when the conditions in Definition 1.4 are not met, t-SNE can still provably visualize at least one cluster in the original data in some cases. As an example, using a more fine-grained analysis, we show that t-SNE computes a partial visualization for data obtained from a mixture of two *concentric* (thus, no mean separation at all!) spherical Gaussians with variances differing by a constant factor.

**Theorem 1.7** (Informal, see Theorem 4.1 for details). *Let $\mathcal{X}$ be generated from an equal-weighted mixture of two Gaussians $\mathcal{N}(0, \sigma_1^2)$ and $\mathcal{N}(0, \sigma_2^2)$ such that $1.5 \leq \sigma_2/\sigma_1 \leq 10$. Then t-SNE with early exaggeration on input $\mathcal{X}$ outputs a $(1 - d^{-\Omega(1)})$-partial visualization of $\mathcal{X}$ where $\mathcal{C}_1$ is $(1 - d^{-\Omega(1)})$-visible.*

## 1.2 Related Work

This paper continues the line of work focused on analyzing gradient descent and related heuristics for non-convex optimization problems, examples of which we have discussed before. Theoretically analyzing t-SNE, in particular, was recently considered in a work of Linderman and Steinerberger (2017) who showed that running t-SNE with early exaggeration causes points from the same cluster to move towards each other (i.e., embedding of any cluster shrinks). As discussed before, however, this does not imply that t-SNE ends up with a visualization as all the clusters could potentially collapse into each other. Another work by Shaham and Steinerberger (2017) derived a theoretical property of SNE, but their result is only nontrivial when the number of clusters is significantly larger than the number of points per cluster, which is an unrealistic assumption.

Mixture models are natural average-case generative models for clusterable data which have been studied as benchmarks for analyzing various clustering algorithms and have a long history of theoretical work. By now, a sequence of results (Dasgupta et al., 2007, 2006; Arora and Kannan, 2005; Vempala and Wang, 2004; Achlioptas and McSherry, 2005; Kannan et al., 2005; Vempala, 2007; Hsu and Kakade, 2013; Ge et al., 2015b; Kalai et al., 2012; Belkin and Sinha, 2010; Kalai et al., 2010; Kothari and Steinhardt, 2017; Hopkins and Li, 2017; Diakonikolas et al., 2017) have identified efficient algorithms for clustering data from such models under various natural assumptions.

## 2 Preliminaries and Notation

We use $\|x\|$ to denote the Euclidean norm of a vector $x$, and use $\langle x, y \rangle$ to denote the standard Euclidean inner product $x^\top y$. We denote by $\|A\|_F$ and $\|A\|_2$ respectively the Frobenius norm and the spectral norm of a matrix $A$. For two real numbers $a, b$ ($b > 0$), we use $a \pm b$ to represent any number in the interval $[a-b, a+b]$. For $x \in \mathbb{R}^s$ and $r > 0$, denote by $\mathcal{B}(x, r)$ the Euclidean ball in $\mathbb{R}^s$ of radius $r$ centered at $x$, i.e., $\mathcal{B}(x, r) := \{x' \in \mathbb{R}^s : \|x - x'\| \le r\}$. For any sets $\mathcal{S}_1, \mathcal{S}_2 \subseteq \mathbb{R}^s$, define $\mathcal{S}_1 + \mathcal{S}_2 := \{a + b : a \in \mathcal{S}_1, b \in \mathcal{S}_2\}$. Denote by $\mathsf{Diam}(\mathcal{S})$ the diameter of a bounded set $\mathcal{S} \subset \mathbb{R}^s$, i.e., $\mathsf{Diam}(\mathcal{S}) := \sup_{x,y \in \mathcal{S}} \|x - y\|$. By $A \gg B$ or $B \ll A$ ($A, B > 0$), we mean $A > cB$ for a large enough constant $c > 0$.

In this paper, we consider a dataset $\mathcal{X} = \{x_1, \ldots, x_n\} \subset \mathbb{R}^d$ with ground-truth clustering $\mathcal{C}_1, \ldots, \mathcal{C}_k$, where $\mathcal{C}_1, \ldots, \mathcal{C}_k$ form a partition of $[n]$. We assume $d$ and $n$ to be sufficiently large, and $n = \mathrm{poly}(d)$. We denote by $\pi : [n] \to [k]$ the function that maps each $i \in [n]$ to the cluster that $x_i$ belongs to, i.e., $i \in \mathcal{C}_{\pi(i)}$. For points $x_i$ and $x_j$ ($i, j \in [n]$) in the input clusterable data, we use $i \sim j$ to mean that $x_i$ and $x_j$ are from the same cluster (i.e., $\pi(i) = \pi(j)$), and $i \not\sim j$ otherwise.

**The t-SNE Algorithm.** The goal of t-SNE is to map $\mathcal{X}$ to a 2 or 3-dimensional dataset $\mathcal{Y} = \{y_1, y_2, \ldots, y_n\}$ that can be visualized in a scatter plot. In this paper we consider visualization in $\mathbb{R}^2$ for simplicity, but our results hold for $\mathbb{R}$, $\mathbb{R}^3$ or any other constant dimension as well.

The t-SNE algorithm starts by computing a joint probability distribution $p_{ij}$ over pairs of points $x_i, x_j$ ($i \ne j$):
$$p_{j|i} = \frac{\exp\left(-\|x_i - x_j\|^2 / 2\tau_i^2\right)}{\sum_{l \in [n] \setminus \{i\}} \exp\left(-\|x_i - x_l\|^2 / 2\tau_i^2\right)}, \qquad p_{ij} = \frac{p_{i|j} + p_{j|i}}{2n}, \tag{1}$$
where $\tau_i$ is a tunable parameter that controls the bandwidth of the Gaussian kernel around point $x_i$. In a two-dimensional map $\mathcal{Y} = \{y_1, \ldots, y_n\} \subset \mathbb{R}^2$, define the affinity $q_{ij}$ between points $y_i$ and $y_j$ ($i \ne j$) as
$$q_{ij} = \frac{(1 + \|y_i - y_j\|^2)^{-1}}{\sum_{l,s \in [n], l \ne s}(1 + \|y_l - y_s\|^2)^{-1}}. \tag{2}$$
Then t-SNE tries to find points $y_i$'s in $\mathbb{R}^2$ that minimize the KL-divergence between $p$ and $q$:
$$f(y_1, \ldots, y_n) := \mathrm{KL}(p\|q) = \sum_{i,j \in [n], i \ne j} p_{ij} \log \frac{p_{ij}}{q_{ij}}.$$
The objective function $f$ is minimized using gradient descent. Its gradient is the following:
$$\frac{\partial f}{\partial y_i} = 4 \sum_{j \in [n] \setminus \{i\}} (p_{ij} - q_{ij}) q_{ij} Z (y_i - y_j), \qquad i \in [n],$$

**Algorithm 1** t-SNE

**Input:** Dataset $\mathcal{X} = \{x_1, \ldots, x_n\} \subset \mathbb{R}^d$, Gaussian bandwidths $\tau_1, \ldots, \tau_n > 0$, exaggeration parameter $\alpha > 0$, step size $h > 0$, number of rounds $T \in \mathbb{N}$

1: Compute $\{p_{ij} : i, j \in [n], i \neq j\}$ using (1)
2: Initialize $y_1^{(0)}, y_2^{(0)}, \ldots, y_n^{(0)}$ i.i.d. from the uniform distribution on $[-0.01, 0.01]^2$
3: **for** $t = 0$ **to** $T - 1$ **do**
4: $\quad Z^{(t)} \leftarrow \sum_{i,j \in [n], i \neq j} \left(1 + \left\|y_i^{(t)} - y_j^{(t)}\right\|^2\right)^{-1}$
5: $\quad q_{ij}^{(t)} \leftarrow \dfrac{\left(1 + \left\|y_i^{(t)} - y_j^{(t)}\right\|^2\right)^{-1}}{Z^{(t)}}, \quad \forall i, j \in [n], i \neq j$
6: $\quad y_i^{(t+1)} \leftarrow y_i^{(t)} + h \sum_{j \in [n] \setminus \{i\}} \left(\alpha p_{ij} - q_{ij}^{(t)}\right) q_{ij}^{(t)} Z^{(t)} \left(y_j^{(t)} - y_i^{(t)}\right), \quad \forall i \in [n]$
7: **end for**

**Output:** 2D embedding $\mathcal{Y}^{(T)} = \left\{y_1^{(T)}, y_2^{(T)}, \ldots, y_n^{(T)}\right\} \subset \mathbb{R}^2$

where $Z = \sum_{l,s \in [n], l \neq s}(1 + \|y_l - y_s\|^2)^{-1}$.

In this paper we consider the *early exaggeration* trick proposed by van der Maaten and Hinton (2008), where all $p_{ij}$'s are multiplied by a factor $\alpha > 1$.[1] Letting the step size in the gradient descent method be $\frac{h}{4}$, we get the following update rule:

$$y_i^{(t+1)} = y_i^{(t)} + h \sum_{j \in [n] \setminus \{i\}} \left(\alpha p_{ij} - q_{ij}^{(t)}\right) q_{ij}^{(t)} Z^{(t)} \left(y_j^{(t)} - y_i^{(t)}\right), \qquad i = 1, \ldots, n. \tag{3}$$

Here $y_i^{(t)} \in \mathbb{R}^2$ is the position of $y_i$ after $t$ iterations. We summarize the t-SNE algorithm in Algorithm 1. Throughout this paper, we let $\left\{y_i^{(0)} : i \in [n]\right\}$ be initialized i.i.d. from the uniform distribution over $[-0.01, 0.01]^2$.

## 3 Full Visualization via t-SNE

In this section, we present our analysis of t-SNE for visualizing spherical and well-separated data. That is, we will assume that for some $\gamma > 0$, we are given a dataset $\mathcal{X} = \{x_1, \ldots, x_n\} \subset \mathbb{R}^d$ with ground-truth clusters $\mathcal{C}_1, \ldots, \mathcal{C}_k$ satisfying Definition 1.4. We will show that t-SNE with early exaggeration on input $\mathcal{X}$ produces a 2-dimensional embedding $\mathcal{Y}$ that is a full visualization of $\mathcal{X}$. Specifically, our main result is the following:

**Theorem 3.1.** *Let* $\mathcal{X} = \{x_1, x_2, \ldots, x_n\} \subset \mathbb{R}^d$ *be* $\gamma$-*spherical and* $\gamma$-*well-separated clusterable data with* $\mathcal{C}_1, \mathcal{C}_2, \ldots, \mathcal{C}_k$ *defining the* $k$ *individual clusters of size at least* $0.1(n/k)$, *where* $k \ll n^{1/5}$. *Choose* $\tau_i^2 = \frac{\gamma}{4} \cdot \min_{j \in [n] \setminus \{i\}} \|x_i - x_j\|^2$ *(*$\forall i \in [n]$*),* $h = 1$, *and any* $\alpha$ *satisfying* $k^2 \sqrt{n} \log n \ll \alpha \ll n$.

*Let* $\mathcal{Y}^{(T)}$ *be the output of t-SNE (Algorithm 1) after* $T = \Theta\left(\frac{n \log n}{\alpha}\right)$ *iterations on input* $\mathcal{X}$ *with the above parameters. Then, with probability at least* 0.99 *over the choice of the initialization,* $\mathcal{Y}^{(T)}$ *is a full visualization of* $\mathcal{X}$.

---
[1] It is suggested in (van der Maaten and Hinton, 2008; van der Maaten, 2014) that early exaggeration (i.e., $\alpha > 1$) is used in the first 50-250 iterations and then the algorithm switches to $\alpha = 1$. In this paper we show that the early exaggeration phase of t-SNE already leads to visualization.

**Remark.** *The original paper (van der Maaten and Hinton, 2008) chooses $\tau_i$ such that the affinities $\{p_{j|i}\}_{j \neq i}$ form a distribution that puts significant mass only on roughly $N$ nearest neighbors of $x_i$ for some parameter $N$ (namely, with perplexity $N$). They suggest choosing $N$ between 5 and 50. In practice $\tau_i$ is found by trial and error (e.g. binary search).*

*A simple analysis shows that the sensible choice of $N$ is roughly the cluster size one expects to see. (For example, when data is generated from a mixture of Gaussians, as $\tau_i$ increases from 0 the perplexity stays low and then undergoes a sharp phase transition when perplexity gets close to cluster size. Thus the behavior is stable to perturbations only at the upper limit.) This is what determines our choice of $\tau_i$ in Theorem 3.1.*

Observe that as $\gamma$ becomes smaller, our analysis shows that t-SNE requires less separation between individual clusters in $\mathcal{X}$ in order to succeed in finding a full visualization of $\mathcal{X}$. By verifying these conditions using standard Gaussian concentration results (see e.g. (Arora and Kannan, 2005)), we immediately obtain the following corollary that t-SNE produces a full visualization for well-separated mixture of isotropic Gaussians. This result holds more generally for some non-isotropic mixtures of log-concave distributions too – for details see Section 3.2.

**Corollary 3.2.** *Let $\mathcal{X} = \{x_1, \ldots, x_n\}$ be generated i.i.d. from a mixture of $k$ Gaussians $\mathcal{N}(\mu_i, I)$ whose means $\mu_1, \mu_2, \ldots, \mu_k$ satisfy $\|\mu_\ell - \mu_{\ell'}\| = \tilde{\Omega}(d^{1/4})$ for any $\ell \neq \ell'$.*

*Let $\mathcal{Y}$ be the output of the t-SNE algorithm with early exaggeration when run on input $\mathcal{X}$ with parameters from Theorem 3.1. Then, with high probability over the draw of $\mathcal{X}$ and the choice of the random initialization, $\mathcal{Y}$ is a full visualization of $\mathcal{X}$.*

In the rest of this section, we establish the above results.

*Proof of Theorem 3.1.* We break the proof of Theorem 3.1 into two parts: (i) Lemma 3.3, which identifies sufficient conditions on the pairwise affinities $p_{ij}$'s that imply that t-SNE outputs a full visualization, and (ii) Lemma 3.4, which shows that the $p_{i,j}$'s computed for $\gamma$-spherical, $\gamma$-well-separated data satisfy the requirements in Lemma 3.3. Combining Lemmas 3.3 and 3.4 gives Theorem 3.1. □

**Lemma 3.3.** *Consider a dataset $\mathcal{X} = \{x_1, x_2, \ldots, x_n\} \subseteq \mathbb{R}^d$ with ground-truth clusters $\mathcal{C}_1, \ldots, \mathcal{C}_k$ satisfying $|\mathcal{C}_\ell| \geq 0.1(n/k)$ for each $\ell \in [k]$ and $k \ll n^{1/5}$. Let $\{p_{ij} : i, j \in [n], i \neq j\}$ be the pairwise affinities in t-SNE computed by (1). Suppose that there exist $\delta, \epsilon, \eta > 0$ such that $\{p_{ij}\}$, $\alpha$ and $h$ in t-SNE (Algorithm 1) satisfy:*

(i) *for any cluster $\ell$, and any point $i \in C_\ell$, we have $\left|\left\{j \in C_\ell : \alpha h p_{ij} \geq \frac{\delta}{|C_\ell|}\right\}\right| \geq \left(\frac{1}{2} + \eta\right)|C_\ell|$;*

(ii) *for any cluster $\ell$, and any point $i \in C_\ell$, we have $\alpha h \sum_{j \in C_\ell \setminus \{i\}} p_{ij} \leq 1$;*

(iii) *for any cluster $\ell$, and any point $i \in C_\ell$, we have $\alpha h \sum_{j \notin C_\ell} p_{ij} + \frac{h}{n} \leq \epsilon$;*

(iv) $\frac{\epsilon \log \frac{1}{\epsilon}}{\delta \eta} \ll \frac{1}{k^2 \sqrt{n}}$.

*Then, with high probability over the choice of the random initialization, the output $\mathcal{Y}^{(T)}$ of t-SNE after $T = \Theta\left(\frac{\log \frac{1}{\epsilon}}{\delta \eta}\right)$ iterations is a full visualization of $\mathcal{X}$.*

**Lemma 3.4.** *Let $\mathcal{X} = \{x_1, x_2, \ldots, x_n\} \subseteq \mathbb{R}^d$ be $\gamma$-spherical and $\gamma$-well-separated clusterable data with $\mathcal{C}_1, \mathcal{C}_2, \ldots, \mathcal{C}_k$ defining the individual clusters such that $|\mathcal{C}_i| \geq 0.1 n/k$ for every $i$. Let $p_{i,j}$'s be the affinities computed by t-SNE (Algorithm 1) with parameters $\tau_i^2 = \frac{\gamma}{4} \cdot \min_{j \in [n] \setminus \{i\}} \|x_i - x_j\|^2$ ($\forall i \in [n]$), $h = 1$, and any $\alpha$ satisfying $k^2 \sqrt{n} \log n \ll \alpha \ll n$.*

*Then, $p_{ij}$'s satisfy (i)-(iv) in Lemma 3.3 with $\delta = \Theta(\alpha/n), \epsilon = 2/n$ and $\eta = 0.01$.*

We prove Lemma 3.4 in Appendix A.1.

## 3.1 Proof of Lemma 3.3

The proof of Lemma 3.3 is naturally divided into two parts. In the first part, we establish that over the course of the iterative updates in the t-SNE algorithm, distances between points in the same cluster decrease (Lemma 3.5). This step is essentially done in the work of Linderman and Steinerberger (2017). We include a full proof in Appendix A.2 for completeness.

**Lemma 3.5** (Shrinkage of clusters). *Under the same setting as Lemma 3.3, after running t-SNE for $T = \Theta\left(\frac{\log \frac{1}{\epsilon}}{\delta \eta}\right)$ rounds, we have* $\mathsf{Diam}\left(\left\{y_i^{(T)} : i \in \mathcal{C}_\ell\right\}\right) = O\left(\frac{\epsilon}{\delta \eta}\right)$ *for all $\ell \in [k]$.*

In the second part, we establish that points in different clusters remain separated in the embedding if the clusters are well-separated in the input data. Concretely, let $\mu_\ell^{(t)} := \frac{1}{|\mathcal{C}_\ell|} \sum_{i \in \mathcal{C}_\ell} y_i^{(t)}$, which is the *centroid* of $\left\{y_i^{(t)} : i \in \mathcal{C}_\ell\right\}$. The following lemma says that the centroids of all clusters will remain separated in the first $O\left(\frac{\log \frac{1}{\epsilon}}{\delta \eta}\right)$ rounds.

**Lemma 3.6** (Separation of clusters). *Under the same setting as Lemma 3.3, if t-SNE is run for $T = O\left(\frac{\log \frac{1}{\epsilon}}{\delta \eta}\right)$ iterations, with high probability we have $\left\|\mu_\ell^{(T)} - \mu_{\ell'}^{(T)}\right\| = \Omega\left(\frac{1}{k^2 \sqrt{n}}\right)$ for all $\ell \neq \ell'$.*

We can finish the proof of Lemma 3.3 using the above two lemmas.

*Proof of Lemma 3.3.* Using Lemmas 3.5 and 3.6, we know that after $T = \Theta\left(\frac{\log \frac{1}{\epsilon}}{\delta \eta}\right)$ iterations, for any $i, j \in [n]$ we have:

- if $i \sim j$, then $\left\|y_i^{(T)} - y_j^{(T)}\right\| \leq \mathsf{Diam}\left(\left\{y_l^{(t)} : l \in \mathcal{C}_{\pi(i)}\right\}\right) = O\left(\frac{\epsilon}{\delta \eta}\right)$;

- if $i \not\sim j$, then $\left\|y_i^{(T)} - y_j^{(T)}\right\| \geq \left\|\mu_{\pi(i)}^{(T)} - \mu_{\pi(j)}^{(T)}\right\| - \left\|y_i^{(T)} - \mu_{\pi(i)}^{(T)}\right\| - \left\|y_j^{(T)} - \mu_{\pi(j)}^{(T)}\right\| \geq \left\|\mu_{\pi(i)}^{(T)} - \mu_{\pi(j)}^{(T)}\right\| - \mathsf{Diam}\left(\left\{y_l^{(t)} : l \in \mathcal{C}_{\pi(i)}\right\}\right) - \mathsf{Diam}\left(\left\{y_l^{(t)} : l \in \mathcal{C}_{\pi(j)}\right\}\right) \geq \Omega\left(\frac{1}{k^2 \sqrt{n}}\right) - O\left(\frac{\epsilon}{\delta \eta}\right) - O\left(\frac{\epsilon}{\delta \eta}\right) = \Omega\left(\frac{1}{k^2 \sqrt{n}}\right) \gg O\left(\frac{\epsilon}{\delta \eta}\right)$.

Thus, in $\mathcal{Y}^{(T)}$, each point is much closer to points from its own cluster than to points from other clusters. Therefore $\mathcal{Y}^{(T)}$ is a full visualization of $\mathcal{X}$ with $\mathcal{C}_1, \ldots, \mathcal{C}_k$ being visible clusters. □

### 3.1.1 Proof of Lemma 3.6

Our idea is to track the centroids of the points coming from the true clusters and show that they remain separated at the end of the algorithm. Towards this, we first show random initialization ensures that the cluster centroids are initially well-separated with high probability.

**Lemma 3.7.** *Suppose $|\mathcal{C}_\ell| \geq 0.1(n/k)$ for all $\ell \in [k]$. If $y_i^{(0)}$'s are generated i.i.d. from the uniform distribution over $[-0.01, 0.01]^2$, then with probability at least $0.99$ we have $\left\|\mu_\ell^{(0)} - \mu_{\ell'}^{(0)}\right\| = \Omega\left(\frac{1}{k^2 \sqrt{n}}\right)$ for all $\ell \neq \ell'$.*

To prove Lemma 3.7, first recall the classical Berry-Esseen theorem.

**Lemma 3.8** (Berry-Esseen theorem (Berry, 1941; Esseen, 1942)). *Suppose that $X_1, X_2, \ldots, X_m$ are i.i.d. random variables with $\mathbb{E}[X_1] = 0$, $\mathbb{E}[X_1^2] = \sigma^2 < \infty$ ($\sigma > 0$) and $\mathbb{E}[|X_1|^3] = \zeta < \infty$. Let $Y_m := \frac{1}{m} \sum_{i=1}^m X_i$, $F_m$ be the cumulative distribution function (CDF) of $\frac{Y_m \sqrt{m}}{\sigma}$, and $\Phi$ be the CDF of the standard* normal distribution $\mathcal{N}(0, 1)$. Then there exists a universal constant $C$ such that for all $x \in \mathbb{R}$ and all $m \in \mathbb{N}$ we have
$$|F_m(x) - \Phi(x)| \leq \frac{C\zeta}{\sigma^3 \sqrt{m}}.$$

*Proof of Lemma 3.7.* We ignore the superscript "$(0)$" for ease of presentation. For a vector $y$, denote by $(y)_1$ its first coordinate. Then it suffices to prove $|(\mu_\ell)_1 - (\mu_{\ell'})_1| = \Omega\left(\frac{1}{k^2\sqrt{n}}\right)$ for all $\ell \neq \ell'$.

Consider a fixed $\ell \in [k]$. Note that $(y_i)_1$'s are i.i.d. with the uniform distribution over $[-0.01, 0.01]$, which clearly has zero mean and finite second and third absolute moments. Since $(\mu_\ell)_1 = \frac{1}{|\mathcal{C}_\ell|} \sum_{i \in \mathcal{C}_\ell} (y_i)_1$, using the Berry-Esseen theorem (Lemma 3.8) we know that
$$|F(x) - \Phi(x)| \leq O\left(1/\sqrt{|\mathcal{C}_\ell|}\right),$$

where $F$ is the CDF of $\frac{(\mu_\ell)_1 \sqrt{|\mathcal{C}_\ell|}}{\sigma}$ ($\sigma$ is the standard deviation of the uniform distribution over $[-0.01, 0.01]$), and $\Phi$ is the CDF of $\mathcal{N}(0, 1)$. It follows that for any fixed $a \in \mathbb{R}$ and $b > 0$, we have

$$\begin{aligned}
\Pr\left[|(\mu_\ell)_1 - a| \leq \frac{b}{k^2\sqrt{|\mathcal{C}_\ell|}}\right] &= \Pr\left[\left|\frac{(\mu_\ell)_1\sqrt{|\mathcal{C}_\ell|}}{\sigma} - \frac{a\sqrt{|\mathcal{C}_\ell|}}{\sigma}\right| \leq \frac{b}{k^2\sigma}\right] \\
&= F\left(\frac{a\sqrt{|\mathcal{C}_\ell|} + b/k^2}{\sigma}\right) - F\left(\frac{a\sqrt{|\mathcal{C}_\ell|} - b/k^2}{\sigma}\right) \\
&\leq \Phi\left(\frac{a\sqrt{|\mathcal{C}_\ell|} + b/k^2}{\sigma}\right) - \Phi\left(\frac{a\sqrt{|\mathcal{C}_\ell|} - b/k^2}{\sigma}\right) + O\left(\frac{1}{\sqrt{|\mathcal{C}_\ell|}}\right) \\
&= \int_{\frac{a\sqrt{|\mathcal{C}_\ell|} - b/k^2}{\sigma}}^{\frac{a\sqrt{|\mathcal{C}_\ell|} + b/k^2}{\sigma}} \frac{1}{\sqrt{2\pi}} e^{-\frac{x^2}{2}} \, dx + O\left(\sqrt{k/n}\right) \\
&\leq \int_{\frac{a\sqrt{|\mathcal{C}_\ell|} - b/k^2}{\sigma}}^{\frac{a\sqrt{|\mathcal{C}_\ell|} + b/k^2}{\sigma}} \frac{1}{\sqrt{2\pi}} \, dx + O\left(\sqrt{k/n}\right) \\
&= \frac{1}{\sqrt{2\pi}} \frac{2b}{k^2\sigma} + O\left(\sqrt{k/n}\right).
\end{aligned}$$

From $k \ll n^{1/5}$ we have $\sqrt{k/n} \ll 1/k^2$. Therefore letting $b$ be a sufficiently small constant, we can ensure

$$\Pr\left[|(\mu_\ell)_1 - a| \leq \frac{b}{k^2\sqrt{|\mathcal{C}_\ell|}}\right] \leq \frac{1}{\sqrt{2\pi}} \frac{2b}{k^2\sigma} + O\left(\sqrt{k/n}\right) \leq \frac{0.01}{k^2} \qquad (4)$$

for any $a \in \mathbb{R}$. For any $\ell' \neq \ell$, because $(\mu_\ell)_1$ and $(\mu_{\ell'})_1$ are independent, we can let $a = (\mu_{\ell'})_1$ in (4), which tells us

$$\Pr\left[|(\mu_\ell)_1 - (\mu_{\ell'})_1| \leq \frac{b}{k^2\sqrt{|\mathcal{C}_\ell|}}\right] \leq \frac{0.01}{k^2}.$$

The above inequality holds for any $\ell, \ell' \in [k]$ ($\ell \neq \ell'$). Taking a union bound over all $\ell$ and $\ell'$, we know that with probability at least 0.99 we have $|(\mu_\ell)_1 - (\mu_{\ell'})_1| \geq \frac{b}{k^2\sqrt{|\mathcal{C}_\ell|}} = \Omega\left(\frac{1}{k^2\sqrt{n}}\right)$ for all $\ell, \ell' \in [k]$ ($\ell \neq \ell'$) simultaneously. $\square$

The following lemma says that the centroid of each cluster will move no more than $\epsilon$ in each of the first $\frac{0.01}{\epsilon}$ iterations.

**Lemma 3.9.** *Under the same setting as Lemma 3.3, for all $t \leq \frac{0.01}{\epsilon}$ and all $\ell \in [k]$ we have $\left\|\mu_\ell^{(t+1)} - \mu_\ell^{(t)}\right\| \leq \epsilon$.*

To prove Lemma 3.9, we need the following technical claim, which is proved in Appendix A.2 (see Claims A.4 and A.5).

**Claim 3.10** (Same as Claim A.5). *Under the same setting as Lemma 3.3, for all $t \leq \frac{0.01}{\epsilon}$, we have $y_i^{(t)} \in [-0.02, 0.02]^2$ and $\left\|\epsilon_i^{(t)}\right\| \leq \epsilon$ for all $i \in [n]$, as well as $\left\|y_i^{(t)} - y_j^{(t)}\right\| \leq 0.06$, $0.9 \leq q_{ij}^{(t)} Z^{(t)} \leq 1$ and $\frac{0.9}{n(n-1)} \leq q_{ij}^{(t)} \leq \frac{1}{0.9 n(n-1)}$ for all $i, j \in [n]$ ($i \neq j$).*

*Proof of Lemma 3.9.* Taking the average of (3) for all $i \in \mathcal{C}_\ell$, we obtain

$$\frac{1}{|\mathcal{C}_\ell|} \sum_{i \in \mathcal{C}_\ell} y_i^{(t+1)} = \frac{1}{|\mathcal{C}_\ell|} \sum_{i \in \mathcal{C}_\ell} y_i^{(t)} + \frac{h}{|\mathcal{C}_\ell|} \sum_{i \in \mathcal{C}_\ell} \sum_{j \neq i} \left(\alpha p_{ij} - q_{ij}^{(t)}\right) q_{ij}^{(t)} Z^{(t)} \left(y_j^{(t)} - y_i^{(t)}\right)$$

$$= \frac{1}{|\mathcal{C}_\ell|} \sum_{i \in \mathcal{C}_\ell} y_i^{(t)} + \frac{h}{|\mathcal{C}_\ell|} \sum_{i \in \mathcal{C}_\ell} \sum_{j \in \mathcal{C}_\ell, j \neq i} \left(\alpha p_{ij} - q_{ij}^{(t)}\right) q_{ij}^{(t)} Z^{(t)} \left(y_j^{(t)} - y_i^{(t)}\right)$$

$$+ \frac{h}{|\mathcal{C}_\ell|} \sum_{i \in \mathcal{C}_\ell} \sum_{j \notin \mathcal{C}_\ell} \left(\alpha p_{ij} - q_{ij}^{(t)}\right) q_{ij}^{(t)} Z^{(t)} \left(y_j^{(t)} - y_i^{(t)}\right)$$

$$= \frac{1}{|\mathcal{C}_\ell|} \sum_{i \in \mathcal{C}_\ell} y_i^{(t)} + \frac{h}{|\mathcal{C}_\ell|} \sum_{i,j \in \mathcal{C}_\ell, i \neq j} \left(\alpha p_{ij} - q_{ij}^{(t)}\right) q_{ij}^{(t)} Z^{(t)} \left(y_j^{(t)} - y_i^{(t)}\right)$$

$$+ \frac{h}{|\mathcal{C}_\ell|} \sum_{i \in \mathcal{C}_\ell} \sum_{j \notin \mathcal{C}_\ell} \left(\alpha p_{ij} - q_{ij}^{(t)}\right) q_{ij}^{(t)} Z^{(t)} \left(y_j^{(t)} - y_i^{(t)}\right)$$

$$= \frac{1}{|\mathcal{C}_\ell|} \sum_{i \in \mathcal{C}_\ell} y_i^{(t)} + \frac{h}{|\mathcal{C}_\ell|} \sum_{i \in \mathcal{C}_\ell} \sum_{j \notin \mathcal{C}_\ell} \left(\alpha p_{ij} - q_{ij}^{(t)}\right) q_{ij}^{(t)} Z^{(t)} \left(y_j^{(t)} - y_i^{(t)}\right).$$

Thus we have

$$\left\|\mu_\ell^{(t+1)} - \mu_\ell^{(t)}\right\| = \left\|\frac{h}{|\mathcal{C}_\ell|} \sum_{i \in \mathcal{C}_\ell} \sum_{j \notin \mathcal{C}_\ell} \left(\alpha p_{ij} - q_{ij}^{(t)}\right) q_{ij}^{(t)} Z^{(t)} \left(y_j^{(t)} - y_i^{(t)}\right)\right\|$$

$$\leq \frac{h}{|\mathcal{C}_\ell|} \sum_{i \in \mathcal{C}_\ell} \sum_{j \notin \mathcal{C}_\ell} \left(\alpha p_{ij} + q_{ij}^{(t)}\right) q_{ij}^{(t)} Z^{(t)} \left\|y_j^{(t)} - y_i^{(t)}\right\|.$$

Since $t \leq \frac{0.01}{\epsilon}$, we can apply Claim 3.10 and get

$$\left\|\mu_\ell^{(t+1)} - \mu_\ell^{(t)}\right\| \leq \frac{h}{|\mathcal{C}_\ell|} \sum_{i \in \mathcal{C}_\ell} \sum_{j \notin \mathcal{C}_\ell} \left(\alpha p_{ij} + \frac{1}{0.9 n(n-1)}\right) \cdot 1 \cdot 0.06$$

$$\leq \frac{1}{|\mathcal{C}_\ell|} \sum_{i \in \mathcal{C}_\ell} \left(\alpha h \sum_{j \notin \mathcal{C}_\ell} p_{ij} + \frac{h}{0.9 n}\right) \cdot 0.06$$

$$\leq \frac{0.06}{|\mathcal{C}_\ell|} \sum_{i \in \mathcal{C}_\ell} \frac{\epsilon}{0.9}$$

$$\leq \epsilon,$$

where we have used $\alpha h \sum_{j \notin \mathcal{C}_\ell} p_{ij} + \frac{h}{n} \leq \epsilon$ for all $i \in \mathcal{C}_\ell$ (condition (iii) in Lemma 3.3). □

Using Lemmas 3.7 and 3.9, we can complete the proof of Lemma 3.6.

*Proof of Lemma 3.6.* Notice that we have $T = O\left(\frac{\log \frac{1}{\epsilon}}{\delta \eta}\right) \ll \frac{1}{k^2 \sqrt{n} \epsilon} < \frac{0.01}{\epsilon}$, where we have used condition (iv) in Lemma 3.3. Hence we can apply Lemma 3.9 for all $t \leq T$.

Lemma 3.9 says that after each iteration every centroid moves by at most $\epsilon$. This means that the distance between any two centroids changes by at most $2\epsilon$. Since with high probability the initial distance between $\mu_\ell^{(0)}$ and $\mu_{\ell'}^{(0)}$ is at least $\Omega\left(\frac{1}{k^2\sqrt{n}}\right)$ (Lemma 3.7), we know that after $T$ rounds we have $\left\|\mu_\ell^{(T)} - \mu_{\ell'}^{(T)}\right\| \geq \Omega\left(\frac{1}{k^2\sqrt{n}}\right) - T \cdot 2\epsilon = \Omega\left(\frac{1}{k^2\sqrt{n}}\right) - O\left(\frac{\epsilon \log \frac{1}{\epsilon}}{\delta \eta}\right) = \Omega\left(\frac{1}{k^2\sqrt{n}}\right)$ with high probability, where the last step is due to condition (iv). □

## 3.2 Visualization of a Mixture of Log-Concave Distributions

We show that Corollary 3.2 (mixture of isotropic Gaussians or isotropic log-concave distributions) can be significantly generalized to mixture of (non-isotropic) log-concave distributions. Namely, we have the following corollary.

**Corollary 3.11.** *Let $\mathcal{X} = \{x_1, \ldots, x_n\} \subset \mathbb{R}^d$ be i.i.d. samples from a mixture of $k$ log-concave distributions with means $\mu_1, \mu_2, \ldots \mu_k$, covariances $\Sigma_1, \Sigma_2, \ldots, \Sigma_k$ and all mixing weights at least $\frac{0.2}{k}$. Let $\sigma_\ell := \sqrt{\|\Sigma_\ell\|_2}$ for every $\ell \in [k]$. Suppose that the radius of each of the $k$ components is equal to $R = \sqrt{\mathrm{tr}(\Sigma_\ell)}$ and suppose $R/\sigma_\ell > d^\eta$ for some constant $\eta > 0$ for every $\ell \in [k]$. Finally, suppose that the means have separation $\|\mu_\ell - \mu_{\ell'}\| \gg Rd^{-\eta/6} \log^{2/3} n$ for all $\ell \neq \ell'$. Let $\mathcal{Y}$ be the output of the t-SNE (Algorithm 1) with the same parameter choices as in Theorem 3.1 when run on $\mathcal{X}$. Then, with probability at least $0.99$ over the choice of the random initialization and the draw of $\mathcal{X}$, $\mathcal{Y}$ is a full visualization of $\mathcal{X}$.*

**Remark.** *For a log-concave distribution, the distance from any sample to the mean concentrates tightly around the radius $R = \sqrt{\mathrm{tr}(\Sigma)}$ where $\Sigma$ is the covariance of the distribution. The mean separation required for t-SNE in our analysis is $Rd^{-\eta/6} \log^{2/3} n$ which is asymptotically smaller than $R$ in high dimensions. Also note that $R/\sigma_\ell = \sqrt{\frac{\mathrm{tr}(\Sigma_\ell)}{\|\Sigma_\ell\|_2}}$ is $\sqrt{d}$ for spherical covariances and the assumption that $R/\sigma_\ell > d^\eta$ essentially says that the distribution should not be low-dimensional, i.e., each of the log-concave components should have nontrivial variances at $d^{\Omega(1)}$ many directions.*

The key component in the proof of Corollary 3.11 is the following distance concentration bound on two samples from two log-concave distributions.

**Lemma 3.12** (Distance concentration for log-concave distributions). *Under the same setting as Corollary 3.11, with high probability, for all $i, j \in [n]$ ($i \neq j$), we have*

$$\|x_i - x_j\|^2 = 2R^2 + \left\|\mu_{\pi(i)} - \mu_{\pi(j)}\right\|^2 \pm O\left(R^2 d^{-\eta/3} \log^{1/3} n\right) \pm O\left(Rd^{-\eta} \log n\right) \left\|\mu_{\pi(i)} - \mu_{\pi(j)}\right\|.$$

The proof of Lemma 3.12 is given in Appendix A.3. Using Lemma 3.12, we can prove Corollary 3.11.

*Proof of Corollary 3.11.* We prove the corollary by showing that the dataset $\mathcal{X}$ satisfies the $\gamma$-sphericalness and $\gamma$-well-separation properties for some $\gamma > 0$ and applying Theorem 3.1. First of all, since each component in the mixture has weight at least $0.2/k$, with high probability we have $|\mathcal{C}_\ell| \geq 0.1(n/k)$ for all $\ell \in [k]$.

Suppose that the high-probability event in Lemma 3.12 happens. Then for any two points $x_i$ and $x_j$ ($i \neq j$), if $i \sim j$, we have

$$\|x_i - x_j\|^2 = 2R^2 \pm O\left(R^2 d^{-\eta/3} \log^{1/3} n\right); \tag{5}$$

if $i \not\sim j$, we have

$$\|x_i - x_j\|^2 = 2R^2 + \|\mu_{\pi(i)} - \mu_{\pi(j)}\|^2 \pm O\left(R^2 d^{-\eta/3} \log^{1/3} n\right) \pm O\left(Rd^{-\eta} \log n\right) \|\mu_{\pi(i)} - \mu_{\pi(j)}\|. \quad (6)$$

Now note that (5) means that each cluster is $\gamma = O\left(d^{-\eta/3}\log^{1/3} n\right)$-spherical, because all the pairwise intra-cluster distances are within $1 \pm \gamma$ of each other. To show $\gamma$-well-separation, it suffices to ensure that the lower bound in (6) is at least $(1 + \gamma \log n)$ times larger than the upper bound in (5), i.e.,

$$2R^2 + \|\mu_{\pi(i)} - \mu_{\pi(j)}\|^2 - O\left(R^2 d^{-\eta/3} \log^{1/3} n\right) - O\left(Rd^{-\eta} \log n\right) \|\mu_{\pi(i)} - \mu_{\pi(j)}\|$$
$$\geq \left(1 + O\left(d^{-\eta/3}\log^{4/3} n\right)\right)\left(2R^2 + O\left(R^2 d^{-\eta/3} \log^{1/3} n\right)\right), \quad \forall i \not\sim i'.$$

This can be ensured when $\|\mu_{\pi(i)} - \mu_{\pi(j)}\| \gg Rd^{-\eta/6}\log^{2/3} n$. This completes the proof. □

### 3.3 Connection to the Noisy Power Method

Our analysis of the t-SNE update rule exploits its close relationship to the update rules in the noisy power method. While it was not made rigorous, similarity with the power method was discussed in the work of Linderman and Steinerberger (2017). Recall that given a matrix $A \in \mathbb{R}^{m \times m}$, the updates in the noisy power method are of the form $z \leftarrow Az + \varepsilon$, where $z \in \mathbb{R}^m$ is the variable for $\varepsilon \in \mathbb{R}^m$ is a (not necessarily random) noise vector. In the t-SNE update rule, if we ignore $q_{i,j}$'s, then we can obtain the following update rule:

$$y_i^{(t+1)} = y_i^{(t)} + \alpha \sum_{j \in [n] \setminus \{i\}} p_{ij}\left(y_j^{(t)} - y_i^{(t)}\right), \quad i = 1, \ldots, n. \quad (7)$$

This can be immediately seen as the (noiseless) iterations in power method for a matrix populated with the pairwise affinities between the input points. It is not difficult to check that the proof of Lemma 3.3 still works if we use the power method updates (7). This implies that for the purpose of computing a full visualization of well-separated $\gamma$-spherical data, t-SNE with appropriately chosen exaggeration parameter $\alpha$ has the same guarantee as the simple power method (7) applied to the matrix of pairwise affinities.

Going further, we recall that the convergence of the power method is determined by the spectral gap of the underlying matrix. Therefore it is natural to try to replace the conditions in Lemma 3.3 with a set of spectral conditions. We have the following lemma which makes this intuition precise. Its proof is similar to our analysis of t-SNE in this section. We give a sketch in Appendix A.4.

**Lemma 3.13.** *Consider a dataset $\mathcal{X} = \{x_1, x_2, \ldots, x_n\} \subseteq \mathbb{R}^d$ with ground-truth clusters $\mathcal{C}_1, \ldots, \mathcal{C}_k$ each with size at least $0.1(n/k)$. Let $\{p_{ij} : i, j \in [n], i \neq j\}$ be the pairwise affinities computed by (1). Given $\{p_{ij}\}$ as well as $\alpha > 0$, for each cluster $\ell \in [k]$ define a $|\mathcal{C}_\ell| \times |\mathcal{C}_\ell|$ matrix $B^\ell = \left(B^\ell_{ij}\right)_{i,j \in \mathcal{C}_\ell}$ as $B^\ell_{ij} = \alpha p_{ij}$ $(i, j \in \mathcal{C}_\ell, i \neq j)$ and $B^\ell_{ii} = 1 - \alpha \sum_{j' \in \mathcal{C}_\ell \setminus \{i\}} p_{ij'}$ $(i \in \mathcal{C}_\ell)$. Let $\Delta, \epsilon > 0$ be two parameters such that the following conditions hold:*

(i) *for any $\ell \in [k]$, we have $\lambda_1(B^\ell) = 1$, and $\left|\lambda_s(B^\ell)\right| \leq 1 - \Delta$ for $s = 2, 3, \ldots, |\mathcal{C}_\ell|$;[2]*

(ii) *for any $i \in [n]$, we have $\alpha \sum_{j \not\sim i} p_{ij} \leq \epsilon$;*

(iii) $\frac{\epsilon\left(\sqrt{n} + \log \frac{1}{\epsilon}\right)}{\Delta} \ll \frac{1}{k^2 \sqrt{n}}$.

*Then if $y_i^{(0)}$'s are generated i.i.d. from the uniform distribution over $[-0.01, 0.01]^2$, after running (7) for $T = \Theta\left(\frac{\log \frac{1}{\epsilon}}{\Delta}\right)$ iterations, $\mathcal{Y}^{(T)} = \left\{y_1^{(T)}, \ldots, y_n^{(T)}\right\}$ will with high probability be a full visualization of $\mathcal{X}$ with visible clusters $\mathcal{C}_1, \ldots, \mathcal{C}_k$.*

---
[2]$\lambda_i(A)$ is the $i$-th largest eigenvalue of a symmetric matrix $A$.

## 4 Partial Visualization via t-SNE

While the full visualization guarantees in Section 3 already demonstrate the power of t-SNE over standard dimensionality reduction techniques, it is natural to ask if the separation conditions on data are necessary for the success of the algorithm. In this section, we investigate this question and show t-SNE outputs a partial visualization (i.e., to visualize at least one cluster, see Definition 1.3) even when the separation condition in Definition 1.4 fails drastically. As an illustrative example, consider the case of data generated according to a mixture of two *concentric* spherical Gaussians $\mathcal{N}(0, \sigma_1^2 I)$ and $\mathcal{N}(0, \sigma_2^2 I)$ in $\mathbb{R}^d$, where $\sigma_2 = 1.5\sigma_1$. Observe that the Gaussians have zero separation between their means! Indeed, for independent samples $x \sim \mathcal{N}(0, \sigma_1^2 I)$ and $y, z \sim \mathcal{N}(0, \sigma_2^2 I)$, standard concentration results tell us that with high probability we have $\|y - z\| > \|y - x\|$, i.e., points from the second (larger variance) Gaussian are closer to points from the first Gaussian than to each other.

Most algorithms for learning mixture of Gaussians fail in such a situation with the exception of distance-based methods such as Arora and Kannan (2005). We show that t-SNE succeeds in visualizing at least one cluster correctly in the above situation.

**Theorem 4.1.** *Suppose $n > 4d$. Let $\mathcal{C}_1, \mathcal{C}_2$ be a partition of $[n]$ such that $|\mathcal{C}_1| = |\mathcal{C}_2| = n/2$. Let $\mathcal{X} = \{x_1, \ldots, x_n\} \subset \mathbb{R}^d$ be generated from the mixture of two Gaussians $\mathcal{N}(0, \sigma_1^2)$ and $\mathcal{N}(0, \sigma_2^2)$ such that $1.5 \leq \frac{\sigma_2}{\sigma_1} \leq 10$,[3] where $x_i$ is generated from $\mathcal{N}(0, \sigma_\ell^2)$ if $i \in \mathcal{C}_\ell$ ($\ell = 1, 2$). Choose $\tau_i^2 = \frac{1}{\sqrt{2}} \cdot \min_{j \in [n] \setminus \{i\}} \|x_i - x_j\|^2$ ($\forall i \in [n]$), $h = 1$, and $\alpha = \rho n$ for a sufficiently small constant $\rho > 0$.*

*Let $\mathcal{Y}^{(T)}$ be the output of t-SNE (Algorithm 1) after $T = \Theta(\log d)$ iterations on input $\mathcal{X}$ with the above parameters. Then, with high probability over the choice of the initialization, $\mathcal{Y}^{(T)}$ is a $(1 - d^{-\Omega(1)})$-partial visualization of $\mathcal{X}$ where $\mathcal{C}_1$ is $(1 - d^{-\Omega(1)})$-visible.*

We prove Theorem 4.1 in Appendix B. In the proof, we describe conditions on clusterable data with $k$ clusters under which t-SNE outputs a partial visualization. Theorem 4.1 follows by verifying these conditions for data generated from a mixture of two concentric Gaussians.

## 5 Conclusion

In this paper, we present a natural rigorous framework for studying the problem of visualization, and give an analysis of a popular heuristic used for visualization, the t-SNE algorithm. Our analysis shows that for clusterable data generated from well-studied generative models, t-SNE provably succeeds in natural parameter regimes where standard linear dimensionality reduction techniques fail.

Nevertheless, our formalization of visualization should be considered a first cut. Indeed, any exhaustive characterization of a good visualization for the human eyes would likely involve psychological and perceptional aspects. From a learning theory perspective, we believe investigating useful notions of visualizations and building a repertoire of analytical and algorithmic tools for the problem is worthy of a thorough further inquiry.

### Acknowledgments

This research was done with support from NSF, ONR, Simons Foundation, Mozilla Research, Schmidt Foundation, DARPA, and SRC.

---

[3] Here 1.5 and 10 can be replaced by any constants greater than 1.

# Appendix

## A Proofs for Section 3

### A.1 Proof of Lemma 3.4

*Proof of Lemma 3.4.* For each $i \in [n]$, define $\mathcal{A}_i := \left\{ j \sim i, j \neq i : \|x_i - x_j\|^2 \leq b_{\pi(i)} \right\}$, where $b_{\pi(i)}$ is the same as in Definition 1.4. We know $|\mathcal{A}_i| \geq 0.51 |\mathcal{C}_{\pi(i)}|$.

We first show that there exist constants $c_1, c_2 > 0$ such that for any $i, j \in [n]$ ($i \neq j$), we have

$$p_{ij} \begin{cases} \leq \frac{c_2}{|\mathcal{C}_{\pi(i)}| \cdot n}, & i \sim j, \\ \in \left[ \frac{c_1}{|\mathcal{C}_{\pi(i)}| \cdot n}, \frac{c_2}{|\mathcal{C}_{\pi(i)}| \cdot n} \right], & j \in \mathcal{A}_i, \\ \leq \frac{1}{n^3}, & i \not\sim j. \end{cases} \quad (8)$$

Define $a_\ell := \frac{b_\ell}{1+\gamma}$ for all $\ell \in [k]$. Consider any fixed $i \in [n]$. Suppose $i \in \mathcal{C}_\ell$. From the $\gamma$-sphericalness and $\gamma$-well-separation properties we know $a_\ell \leq \min_{j \in [n] \setminus \{i\}} \|x_i - x_j\|^2 \leq b_\ell$, and thus $\frac{\gamma}{4} a_\ell \leq \tau_i^2 \leq \frac{\gamma}{4} b_\ell$. Let $N_i = \sum_{j \in [n] \setminus \{i\}} \exp\left(-\|x_i - x_j\|^2 / 2\tau_i^2\right)$. Recall that from (1) we have $p_{j|i} = \frac{\exp(-\|x_i - x_j\|^2 / 2\tau_i^2)}{N_i}$ for all $j \in [n] \setminus \{i\}$. Then we have

$$\begin{aligned} p_{j|i} &\leq \frac{\exp(-a_\ell / 2\tau_i^2)}{N_i}, & \forall j \sim i, \\ \frac{\exp(-b_\ell / 2\tau_i^2)}{N_i} &\leq p_{j|i} \leq \frac{\exp(-a_\ell / 2\tau_i^2)}{N_i}, & \forall j \in \mathcal{A}_i, \\ p_{j|i} &\leq \frac{\exp(-(1 + \gamma \log n) b_\ell / 2\tau_i^2)}{N_i}, & \forall j \not\sim i. \end{aligned} \quad (9)$$

Note that we have

$$\frac{\exp(-a_\ell / 2\tau_i^2)}{\exp(-b_\ell / 2\tau_i^2)} = \exp\left(\frac{b_\ell - a_\ell}{2\tau_i^2}\right) = \exp\left(\frac{\gamma a_\ell}{2\tau_i^2}\right) \leq \exp\left(\frac{\gamma a_\ell}{\frac{\gamma}{2} a_\ell}\right) = e^2$$

and

$$\frac{\exp(-b_\ell / 2\tau_i^2)}{\exp(-(1 + \gamma \log n) b_\ell / 2\tau_i^2)} = \exp\left(\frac{\gamma \log n \cdot b_\ell}{2\tau_i^2}\right) \geq \exp\left(\frac{\gamma \log n \cdot b_\ell}{\frac{\gamma}{2} b_\ell}\right) = n^2.$$

As a consequence, letting $f_i := \frac{\exp(-b_\ell / 2\tau_i^2)}{N_i}$, we know that (9) implies

$$\begin{aligned} p_{j|i} &\leq O(f_i), & \forall j \sim i, \\ f_i &\leq p_{j|i} \leq O(f_i), & \forall j \in \mathcal{A}_i, \\ p_{j|i} &\leq \frac{f_i}{n^2}, & \forall j \not\sim i. \end{aligned} \quad (10)$$

Since $\sum_{j \in [n] \setminus \{i\}} p_{j|i} = 1$, from (10) we have

$$1 = \sum_{j \in [n] \setminus \{i\}} p_{j|i} = \sum_{j \sim i, j \neq i} p_{j|i} + \sum_{j \not\sim i} p_{j|i} \leq \sum_{j \sim i, j \neq i} O(f_i) + \sum_{j \not\sim i} \frac{f_i}{n^2} \leq O(|\mathcal{C}_\ell| f_i) + \frac{f_i}{n} = O(|\mathcal{C}_\ell| f_i)$$

and

$$1 = \sum_{j \in [n] \setminus \{i\}} p_{j|i} \geq \sum_{j \in \mathcal{A}_i} p_{j|i} \geq |\mathcal{A}_i| f_i \geq \frac{1}{2} |\mathcal{C}_\ell| f_i.$$

Thus we have $f_i = \Theta\left(\frac{1}{|\mathcal{C}_\ell|}\right)$. Then (10) leads to

$$p_{j|i} \begin{cases} = O\left(\frac{1}{|\mathcal{C}_\ell|}\right), & \forall j \sim i, \\ = \Theta\left(\frac{1}{|\mathcal{C}_\ell|}\right), & \forall j \in \mathcal{A}_i, \\ \leq \frac{1}{n^2}, & \forall j \not\sim i. \end{cases}$$

Plugging this into (1), we obtain the desired bounds on $p_{ij}$'s:

$$p_{ij} \begin{cases} = O\left(\frac{1}{|\mathcal{C}_\ell| \cdot n}\right), & i \sim j, \\ = \Theta\left(\frac{1}{|\mathcal{C}_\ell| \cdot n}\right), & j \in \mathcal{A}_i, \\ \leq \frac{1}{n^3}, & i \not\sim j. \end{cases}$$

Therefore, (8) is proved.

Next, based on (8), we verify that the four conditions in Lemma 3.3 are all satisfied with parameters $\delta = \frac{c_1 \alpha}{n}$, $\epsilon = \frac{2}{n}$ and $\eta = 0.01$. Consider any $i \in [n]$.

- For all $j \in \mathcal{A}_i$ we have $\alpha h p_{ij} \geq \frac{\alpha c_1}{|\mathcal{C}_{\pi(i)}| \cdot n} = \frac{\delta}{|\mathcal{C}_{\pi(i)}|}$. Since $|\mathcal{A}_i| \geq \left(\frac{1}{2} + \eta\right) |\mathcal{C}_{\pi(i)}|$, we have verified condition (i).

- We have $\alpha h \sum_{j \sim i, j \neq i} p_{ij} \leq \alpha |\mathcal{C}_{\pi(i)}| \cdot \frac{c_2}{|\mathcal{C}_{\pi(i)}| \cdot n} = \frac{c_2 \alpha}{n} \leq 1$, where we have used $\alpha \ll n$. Hence (ii) is verified.

- We have $\alpha h \sum_{j \not\sim i} p_{ij} + \frac{h}{n} \leq \alpha n \cdot \frac{1}{n^3} + \frac{1}{n} \leq \frac{\rho}{n} + \frac{1}{n} \leq \frac{2}{n} = \epsilon$, which verifies (iii).

- Finally, we have $\frac{\epsilon \log \frac{1}{\epsilon}}{\delta \eta} = \frac{\frac{2}{n} \log \frac{n}{2}}{\frac{c_1 \alpha}{n} \cdot 0.01} = \Omega\left(\frac{\log n}{\alpha}\right) \ll \frac{1}{k^2 \sqrt{n}}$. Here we have used $\alpha \gg k^2 \sqrt{n} \log n$. Hence (iv) is satisfied.

Therefore we have finished the proof of Lemma 3.4. $\square$

## A.2 Proof of Lemma 3.5

**Lemma A.1** (Lemma 1 in (Linderman and Steinerberger, 2017)). *Let $z_1, \ldots, z_m \in \mathbb{R}^s$ be evolving as the following dynamic system:*

$$z_i^{(t+1)} = \sum_{j=1}^m \lambda_{ij}^{(t)} z_j^{(t)} + \epsilon_i^{(t)}, \qquad i \in [m],\ t = 0, 1, 2, \ldots \tag{11}$$

*where $z_i^{(t)}$ is the position of $z_i$ at time $t$. Denote by $\mathsf{Conv}^{(t)}$ the convex hull of $z_1^{(t)}, \ldots, z_m^{(t)}$, and let $D^{(t)} := \mathsf{Diam}\left(\mathsf{Conv}^{(t)}\right)$. Suppose at time $t$ we have*

- $\lambda_{ij}^{(t)} \geq 0$ $(\forall i, j \in [m])$, $\sum_{j=1}^m \lambda_{ij}^{(t)} = 1$ $(\forall i \in [m])$;

- $\left\|\epsilon_i^{(t)}\right\| \leq \epsilon'$ $(\forall i \in [m])$.

Then we have $\mathsf{Conv}^{(t+1)} \subseteq \mathsf{Conv}^{(t)} + \mathcal{B}(0, \epsilon')$.

**Lemma A.2** (Lemma 2 in (Linderman and Steinerberger, 2017)). *Under the same setting as Lemma A.1, if furthermore there exist $\delta', \eta' > 0$ such that:*

$$\left|\left\{j : \lambda_{ij}^{(t)} \geq \delta'\right\}\right| \geq \left(\frac{1}{2} + \eta'\right) m, \qquad \forall i \in [m],$$

*then we have* $D^{(t+1)} \leq (1 - m\delta'\eta'/2)D^{(t)} + 2\epsilon'$.

**Remark.** *In fact, Lemma 2 in (Linderman and Steinerberger, 2017) only concerns the case $\eta' = \frac{1}{2}$, but their proof can be directly generalized to any $\eta' \in [0, \frac{1}{2}]$.*

Lemma A.2 implies that the diameter of the points in the dynamic system is shrinking exponentially until its size becomes $O\left(\frac{\epsilon'}{m\delta'\eta'}\right)$; afterwards it remains to be $O\left(\frac{\epsilon'}{m\delta'\eta'}\right)$. Namely, we have:

**Corollary A.3.** *Under the same setting as Lemma A.2, we have: (1) if $D^{(t)} \geq \frac{5\epsilon'}{m\delta'\eta'}$, then $D^{(t+1)} \leq (1 - m\delta'\eta'/10)D^{(t)}$; (2) if $D^{(t)} \leq \frac{5\epsilon'}{m\delta'\eta'}$, then $D^{(t+1)} \leq \frac{5\epsilon'}{m\delta'\eta'}$.*

*Proof.* If $D^{(t)} \geq \frac{5\epsilon'}{m\delta'\eta'}$, we have $\epsilon' \leq \frac{m\delta'\eta'}{5}D^{(t)}$ and then using Lemma A.2 we have $D^{(t+1)} \leq (1 - m\delta'\eta'/2)D^{(t)} + 2\epsilon' \leq (1 - m\delta'\eta'/2)D^{(t)} + \frac{2m\delta'\eta'}{5}D^{(t)} = (1 - m\delta'\eta'/10)D^{(t)}$.

If $D^{(t)} \leq \frac{5\epsilon'}{m\delta'\eta'}$, from Lemma A.2 we have $D^{(t+1)} \leq (1 - m\delta'\eta'/2)D^{(t)} + 2\epsilon' \leq (1 - m\delta'\eta'/2)\frac{5\epsilon'}{m\delta'\eta'} + 2\epsilon' = \frac{5\epsilon'}{m\delta'\eta'} - \frac{1}{2}\epsilon' \leq \frac{5\epsilon'}{m\delta'\eta'}$. □

*Proof of Lemma 3.5.* We rewrite the evolution of points in $\left\{y_i^{(t)} : i \in \mathcal{C}_\ell\right\}$ (for each $\ell \in [k]$) in the form of the dynamic system (11):

$$\begin{aligned}
y_i^{(t+1)} &= y_i^{(t)} + \alpha h \sum_{j \sim i, j \neq i} p_{ij} q_{ij}^{(t)} Z^{(t)} \left(y_j^{(t)} - y_i^{(t)}\right) \\
&\quad + \alpha h \sum_{j \nsim i} p_{ij} q_{ij}^{(t)} Z^{(t)} \left(y_j^{(t)} - y_i^{(t)}\right) - h \sum_{j \neq i} \left(q_{ij}^{(t)}\right)^2 Z^{(t)} \left(y_j^{(t)} - y_i^{(t)}\right) \\
&= \sum_{j \in \mathcal{C}_\ell} \lambda_{ij}^{(t)} y_j^{(t)} + \epsilon_i^{(t)}, \qquad i \in \mathcal{C}_\ell,
\end{aligned} \quad (12)$$

where

$$\begin{aligned}
\lambda_{ij}^{(t)} &:= \alpha h p_{ij} q_{ij}^{(t)} Z^{(t)}, \qquad i \sim j, i \neq j, \\
\lambda_{ii}^{(t)} &:= 1 - \sum_{j \sim i, j \neq i} \lambda_{ij}^{(t)}, \qquad i \in [n], \\
\epsilon_i^{(t)} &:= \alpha h \sum_{j \nsim i} p_{ij} q_{ij}^{(t)} Z^{(t)} \left(y_j^{(t)} - y_i^{(t)}\right) - h \sum_{j \neq i} \left(q_{ij}^{(t)}\right)^2 Z^{(t)} \left(y_j^{(t)} - y_i^{(t)}\right), \qquad i \in [n].
\end{aligned}$$

We can verify $\lambda_{ij}^{(t)} \geq 0$ for all $i \sim j, i \neq j$, and

$$\lambda_{ii}^{(t)} = 1 - \sum_{j \sim i, j \neq i} \alpha h p_{ij} q_{ij}^{(t)} Z^{(t)} = 1 - \sum_{j \sim i, j \neq i} \alpha h p_{ij} \left(1 + \left\|y_i^{(t)} - y_j^{(t)}\right\|^2\right)^{-1}$$

$$\geq 1 - \sum_{j\sim i, j\neq i} \alpha h p_{ij} \geq 0, \qquad \forall i \in [n],$$

where the last inequality is due to (iii) in Lemma 3.3.

Next, we will use Lemma A.1 to prove the following claim.

**Claim A.4.** *Under the same setting as Lemma 3.3, for all $t \leq \frac{0.01}{\epsilon}$, we have $y_i^{(t)} \in [-0.02, 0.02]^2$ and $\left\|\epsilon_i^{(t)}\right\| \leq \epsilon$ for all $i \in [n]$.*

*Proof.* We prove the claim by proving the following two statements. Throughout the proof we always consider $t \leq \frac{0.01}{\epsilon}$.

(I) If $y_i^{(t)} \in [-0.02, 0.02]^2$ for all $i \in [n]$, then $\left\|\epsilon_i^{(t)}\right\| \leq \epsilon$ for all $i \in [n]$.

(II) If $\left\|\epsilon_i^{(t')}\right\| \leq \epsilon$ for all $i \in [n]$ and all $t' \in \{0, 1, \ldots, t-1\}$, then $y_i^{(t)} \in [-0.02, 0.02]^2$ for all $i \in [n]$.

Since $y_i^{(0)} \in [-0.01, 0.01]^2 \subset [-0.02, 0.02]^2$, it is easy to see that the claim can be proved by repeatedly applying (I) and (II).

First we show (I). Since we have $y_i^{(t)} \in [-0.02, 0.02]^2$ for all $i \in [n]$, we have

$$\left\|y_i^{(t)} - y_j^{(t)}\right\| \leq 0.06, \tag{13}$$

and

$$0.9 \leq q_{ij}^{(t)} Z^{(t)} = \left(1 + \left\|y_i^{(t)} - y_j^{(t)}\right\|^2\right)^{-1} \leq 1, \qquad \forall i, j \in [n], i \neq j. \tag{14}$$

This implies

$$0.9n(n-1) \leq Z^{(t)} = \sum_{i,j \in [n], i\neq j} \left(1 + \left\|y_i^{(t)} - y_j^{(t)}\right\|^2\right)^{-1} \leq n(n-1),$$

and thus

$$\frac{0.9}{n(n-1)} \leq q_{ij}^{(t)} = \frac{\left(1 + \left\|y_i^{(t)} - y_j^{(t)}\right\|^2\right)^{-1}}{Z^{(t)}} \leq \frac{1}{0.9n(n-1)}, \qquad \forall i, j \in [n], i \neq j. \tag{15}$$

Using (13), (14) and (15), we have

$$\begin{aligned}
\left\|\epsilon_i^{(t)}\right\| &\leq \left\|\alpha h \sum_{j\not\sim i} p_{ij} q_{ij}^{(t)} Z^{(t)} \left(y_j^{(t)} - y_i^{(t)}\right)\right\| + \left\|h \sum_{j\neq i} \left(q_{ij}^{(t)}\right)^2 Z^{(t)} \left(y_j^{(t)} - y_i^{(t)}\right)\right\| \\
&\leq \alpha h \sum_{j\not\sim i} p_{ij} q_{ij}^{(t)} Z^{(t)} \left\|y_j^{(t)} - y_i^{(t)}\right\| + h \sum_{j\neq i} q_{ij}^{(t)} \cdot q_{ij}^{(t)} Z^{(t)} \left\|y_j^{(t)} - y_i^{(t)}\right\| \\
&\leq \alpha h \sum_{j\not\sim i} p_{ij} \cdot 1 \cdot 0.6 + h \sum_{j\neq i} \frac{1}{0.9n(n-1)} \cdot 1 \cdot 0.6 \\
&\leq 0.6\alpha h \sum_{j\not\sim i} p_{ij} + 0.7 \cdot \frac{h}{n} \\
&\leq \epsilon, \qquad \forall i \in [n],
\end{aligned} \tag{16}$$

where the last inequality is due to (iii) in Lemma 3.3. This proves (I).

To prove (II), since we have $\left\|\epsilon_i^{(t')}\right\| \leq \epsilon, \forall i \in [n], \forall t' \in [t-1]$, we can repeatedly apply Lemma A.1 and get

$$\begin{aligned}
\mathsf{Conv}\left(\left\{y_i^{(t)} : i \in [n]\right\}\right) &\subseteq \mathsf{Conv}\left(\left\{y_i^{(t-1)} : i \in [n]\right\}\right) + \mathcal{B}(0, \epsilon) \\
&\subseteq \mathsf{Conv}\left(\left\{y_i^{(t-2)} : i \in [n]\right\}\right) + \mathcal{B}(0, 2\epsilon) \\
&\subseteq \cdots \\
&\subseteq \mathsf{Conv}\left(\left\{y_i^{(0)} : i \in [n]\right\}\right) + \mathcal{B}(0, t\epsilon) \\
&\subseteq [-0.01, 0.01]^2 + \mathcal{B}(0, 0.01) \\
&\subseteq [-0.02, 0.02]^2.
\end{aligned}$$

This proves (II). Therefore we have proved Claim A.4. $\square$

In fact, in the above proof we have already shown the following stronger claim:

**Claim A.5.** *Under the same setting as Lemma 3.3, for all $t \leq \frac{0.01}{\epsilon}$, we have $y_i^{(t)} \in [-0.02, 0.02]^2$ and $\left\|\epsilon_i^{(t)}\right\| \leq \epsilon$ for all $i \in [n]$, as well as $\left\|y_i^{(t)} - y_j^{(t)}\right\| \leq 0.06$, $0.9 \leq q_{ij}^{(t)} Z^{(t)} \leq 1$ and $\frac{0.9}{n(n-1)} \leq q_{ij}^{(t)} \leq \frac{1}{0.9n(n-1)}$ for all $i, j \in [n]$ ($i \neq j$).*

Now we return to the proof of Lemma 3.5. We consider any time $t \leq \frac{0.01}{\epsilon}$. Consider any cluster $\mathcal{C}_\ell$ ($\ell \in [k]$) and any $i, j \in \mathcal{C}_\ell$ ($i \neq j$). Using Claim A.5 we have:

$$\alpha h p_{ij} \geq \frac{\delta}{|\mathcal{C}_\ell|} \implies \lambda_{ij}^{(t)} = \alpha h p_{ij} q_{ij}^{(t)} Z^{(t)} \geq \frac{0.9\delta}{|\mathcal{C}_\ell|}.$$

Then from condition (i) in Lemma 3.3 we know $\left|\left\{j \in \mathcal{C}_\ell : \lambda_{ij}^{(t)} \geq \frac{0.9\delta}{|\mathcal{C}_\ell|}\right\}\right| \geq \left(\frac{1}{2} + \eta\right)|\mathcal{C}_\ell|$ for all $i \in \mathcal{C}_\ell$. Recall that we also have $\left\|\epsilon_i^{(t)}\right\| \leq \epsilon$ ($\forall i \in [n]$) according to Claim A.4. Hence the dynamic system (12) satisfy all the conditions in Lemma A.2 and Corollary A.3. Using Corollary A.3, we know $\mathsf{Diam}\left(\left\{y_i^{(T)} : i \in \mathcal{C}_\ell\right\}\right) = O\left(\frac{\epsilon}{|\mathcal{C}_\ell| \cdot \frac{0.9\delta}{|\mathcal{C}_\ell|} \cdot \eta}\right) = O\left(\frac{\epsilon}{\delta\eta}\right)$ as long as $T = \Omega\left(\frac{\log \frac{\epsilon}{\delta\eta}}{\log(1-0.9\delta\eta/10)}\right) = \Omega\left(\frac{\log \frac{\epsilon}{\delta\eta}}{\delta\eta}\right)$. This can be satisfied if $T = \Theta\left(\frac{\log \frac{1}{\epsilon}}{\delta\eta}\right)$ since $\delta\eta = O(1)$. Notice that we still need to check $T \leq \frac{0.01}{\epsilon}$ since otherwise we cannot use Claim A.4 - this can be checked by observing $T = \Theta\left(\frac{\log \frac{1}{\epsilon}}{\delta\eta}\right) \ll \frac{1}{k^2\sqrt{n}\epsilon} < \frac{0.01}{\epsilon}$, where we have used (iv) in Lemma 3.3. Therefore, we have proved Lemma 3.5. $\square$

### A.3 Proof of Lemma 3.12

We first invoke the "thin-shell" property of log-concave distributions:

**Lemma A.6** (Guédon and Milman (2011)). *Suppose that $x$ is drawn from a log-concave distribution in $\mathbb{R}^d$ with mean $\mu$ and covariance $\Sigma$, and $A = \Sigma^{1/2}$.[4] Then:*

$$\Pr\left[\left|\|x - \mu\| - \|A\|_F\right| \geq \epsilon\|A\|_F\right] \leq C \exp\left(-C' \frac{\|A\|_F}{\|A\|_2} \min\{\epsilon, \epsilon^3\}\right), \qquad \forall \epsilon \geq 0,$$

*where $C$ and $C'$ are universal constants.*

---
[4] For a positive semidefinite matrix $\Sigma$, $A = \Sigma^{1/2}$ is a positive semidefinite matrix such that $\Sigma = A^2$.

The following lemma is a corollary of Lemma A.6.

**Lemma A.7.** *Suppose that $x$ is drawn from a log-concave distribution in $\mathbb{R}^d$ with mean $\mu$ and covariance $\Sigma$, and $A = \Sigma^{1/2}$. Then for any $\delta \in (0, 1)$ and any $y \in \mathbb{R}^d$, with probability at least $1 - \delta$ we have $|\langle x - \mu, y \rangle| \leq O\left(\log \frac{1}{\delta}\right) \|Ay\|$.*

*Proof.* Since log-concavity is preserved under linear transformations (Prékopa, 1973), we know that $\langle x - \mu, y \rangle$ also follows a log-concave distribution. Furthermore, we have $\mathbb{E}[\langle x - \mu, y \rangle] = y^\top \mathbb{E}[x - \mu] = 0$ and $\mathbb{E}[\langle x - \mu, y \rangle^2] = y^\top \mathbb{E}[(x - \mu)(x - \mu)^\top] y = y^\top \Sigma y = \|Ay\|^2$. Then invoking Lemma A.6 (with dimension being 1) and letting $\epsilon = \Theta(\log \frac{1}{\delta})$, we know that with probability at least $1 - \delta$, we have $|\langle x - \mu, y \rangle| \leq (1 + \epsilon) \|Ay\| = O(\log \frac{1}{\delta}) \|Ay\|$. This completes the proof. □

We can now prove Lemma 3.12 using Lemmas A.6 and A.7.

*Proof of Lemma 3.12.* Denote $A_\ell := \Sigma_\ell^{1/2}$. Suppose $\pi(i) = \ell$ and $\pi(j) = \ell'$. We have

$$\|x_i - x_j\|^2 = \|(x_i - \mu_\ell) + (\mu_\ell - \mu_{\ell'}) + (\mu_{\ell'} - x_j)\|^2$$
$$= \|x_i - \mu_\ell\|^2 + \|x_j - \mu_{\ell'}\|^2 + \|\mu_\ell - \mu_{\ell'}\|^2 + 2\langle x_i - \mu_\ell, \mu_\ell - \mu_{\ell'}\rangle$$
$$- 2\langle x_j - \mu_{\ell'}, \mu_\ell - \mu_{\ell'}\rangle - 2\langle x_i - \mu_\ell, x_j - \mu_{\ell'}\rangle.$$

We can bound the first two terms using Lemma A.6: choosing $\epsilon^3 = \Theta\left(\frac{\sigma_\ell \log n}{R}\right) < 1$, we know that with probability at least $1 - n^{-10}$ we have $\|x - \mu\| = (1 \pm \epsilon) R = R \pm O\left(R^{2/3} \sigma_\ell^{1/3} \log^{1/3} n\right) = R \pm O\left(R d^{-\eta/3} \log^{1/3} n\right)$; $\|x' - \mu'\|$ can be bounded in the same way. The inner product terms can be bounded using Lemma A.7. (Note that $x_i$ and $x_j$ are independent, so $\langle x_i - \mu_\ell, x_j - \mu_{\ell'}\rangle$ can still be bounded using Lemma A.7 by conditioning on $x_i$ or $x_j$.) Therefore we have

$$\|x_i - x_j\|^2 = R^2 \pm O\left(R^2 d^{-\eta/3} \log^{1/3} n\right) + R^2 \pm O\left(R^2 d^{-\eta/3} \log^{1/3} n\right) + \|\mu_\ell - \mu_{\ell'}\|^2$$
$$\pm O(\log n) \|A_\ell(\mu_\ell - \mu_{\ell'})\| \pm O(\log n) \|A_{\ell'}(\mu_\ell - \mu_{\ell'})\| \pm O(\log n) \|A_{\ell'}(x_i - \mu_\ell)\|$$
$$= 2R^2 \pm O\left(R^2 d^{-\eta/3} \log^{1/3} n\right) + \|\mu_\ell - \mu_{\ell'}\|^2 \pm O(\log n) \sigma_\ell \|\mu_\ell - \mu_{\ell'}\|$$
$$\pm O(\log n) \sigma_{\ell'} \|\mu_\ell - \mu_{\ell'}\| \pm O(R \sigma_{\ell'} \log n)$$
$$= 2R^2 \pm O\left(R^2 d^{-\eta/3} \log^{1/3} n\right) + \|\mu_\ell - \mu_{\ell'}\|^2 \pm O(R d^{-\eta} \log n) \|\mu_\ell - \mu_{\ell'}\|$$
$$\pm O\left(R^2 d^{-\eta} \log n\right)$$
$$= 2R^2 \pm O\left(R^2 d^{-\eta/3} \log^{1/3} n\right) + \|\mu_\ell - \mu_{\ell'}\|^2 \pm O(R d^{-\eta} \log n) \|\mu_\ell - \mu_{\ell'}\|$$

with probability at least $1 - n^{-3}$. Taking a union bound over all pairs $i, j$, we prove the lemma. □

### A.4 Proof Sketch of Lemma 3.13

*Proof Sketch of Lemma 3.13.* For simplicity, in this proof we assume $y_i^{(t)}$'s are scalars. The result holds similarly for $y_i^{(t)} \in \mathbb{R}^2$ by applying the same proof on the two dimensions. We let $y^{(t)}$ be the vector in $\mathbb{R}^n$ whose $i$-th coordinate is $y_i^{(t)}$.

We rewrite the update rule (7) as follows:

$$y_i^{(t+1)} = y_i^{(t)} + \alpha \sum_{j \sim i, j \neq i} p_{ij}\left(y_j^{(t)} - y_i^{(t)}\right) + \alpha \sum_{j \not\sim i} p_{ij}\left(y_j^{(t)} - y_i^{(t)}\right)$$
$$= \sum_{j \in \mathcal{C}_\ell} B_{ij}^\ell y_j^{(t)} + \varepsilon_i^{(t)}, \qquad i \in \mathcal{C}_\ell, \quad (17)$$

where $\varepsilon_i^{(t)} := \alpha \sum_{j \not\sim i} p_{ij}\left(y_j^{(t)} - y_i^{(t)}\right)$.

Let $z^{(t)}$ be the restriction of $y^{(t)}$ on coordinates from $\mathcal{C}_\ell$. Then (17) indicates $z^{(t+1)} = B^\ell z^{(t)} + \varepsilon^{(t)}$, where $\varepsilon^{(t)} \in \mathbb{R}^{\mathcal{C}_\ell}$ is the vector consisting of $\varepsilon_i^{(t)}$'s for all $i \in \mathcal{C}_\ell$. Therefore we arrive at the noisy power method, and its standard analysis tells us that $z^{(t)}$ will converge to the scaled top eigenvector of $B^\ell$ (since we have assumed an eigengap for $B^\ell$), assuming the noises are bounded, which is the case in our setting. Note that the top eigenvector of $B^\ell$ is the all-1 vector, because the all-1 vector is an eigenvector of $B^\ell$ corresponding to eigenvalue 1, and 1 is by assumption the leading eigenvalue of $B^\ell$. The fact that $z^{(t)}$ converges to the scaled all-1 vector means that $y_i^{(t)}$'s ($i \in \mathcal{C}_\ell$) converge to the same value as $t$ grows, i.e., the cluster shrinks in the embedding. This holds for every cluster $\mathcal{C}_\ell$.

On the other hand, one can still prove the same separation bound in Lemma 3.6 with essentially the same proof we presented. Then Lemma 3.13 follows by combining the shrinkage and separation results. $\square$

## B Proofs for Section 4

The main purpose of this section is to prove Theorem 4.1. We will prove a more general result for $k$ clusters (Theorem B.4) and show that this result applies to our example of two concentric Gaussians in Theorem 4.1.

**Definition B.1** (Balanced, $\gamma$-regular data). *Given a dataset $\mathcal{X} = \{x_1, \ldots, x_n\} \subset \mathbb{R}^d$ with ground-truth clusters $\mathcal{C}_1, \ldots, \mathcal{C}_k$, we say that $\mathcal{X}$ is balanced and $\gamma$-regular if there exist positive parameters $\{R_{\ell,\ell'} : \ell, \ell' \in [k]\}$ such that $R_{\ell,\ell'} = R_{\ell',\ell}$ and:*

1. **Balanced:** $|\mathcal{C}_\ell| = \Theta(n/k)$ *for all* $\ell \in [k]$.
2. **$\gamma$-regular:** *for any $i, j \in [n]$ ($i \neq j$), we have $\|x_i - x_j\|^2 = (1 \pm \gamma) R^2_{\pi(i), \pi(j)}$;*
3. $\frac{R_{\ell_1,\ell_2}}{R_{\ell_3,\ell_4}} \leq 10$ *for all* $\ell_1, \ell_2, \ell_3, \ell_4 \in [k]$.

**Remark.** *The $\gamma$-regularity condition is stronger than $\gamma$-sphericalness in Definition 1.4, but Lemma 3.12 implies that it can be satisfied by a large class of mixtures of non-isotropic log-concave distributions with $\gamma = O(d^{-\eta})$ for some $\eta > 0$. For mixtures of spherical Gaussians, it is easy to see that they satisfy $O\left(\sqrt{\frac{\log n}{d}}\right)$-regularity.*

**Fact B.2.** *Consider a balanced $O(d^{-\eta})$-regular ($\eta > 0$ is a constant) dataset $\mathcal{X} = \{x_1, \ldots, x_n\} \subset \mathbb{R}^d$. Suppose we choose $\tau_i = \frac{1}{\sqrt{2}} \min_{j \neq i} \|x_i - x_j\|$ ($\forall i \in [n]$) in t-SNE (Algorithm 1) and compute pairwise affinity $p_{ij}$'s from (1). Then we have $p_{ij} = \frac{c_{\pi(i),\pi(j)}}{n^2} \pm O\left(\frac{1}{n^2 d^\eta}\right)$ for all $i, j \in [n]$ ($i \neq j$), where $c_{\ell,\ell'} = \Theta(1)$ is a constant defined for each pair $\ell, \ell' \in [k]$ ($c_{\ell,\ell'} = c_{\ell',\ell}$).*

*Proof.* From $\gamma$-regularity we know that $2\tau_i^2 = \left(\min_{\ell \in [k]} R^2_{\pi(i),\ell}\right) \cdot (1 \pm O(d^{-\eta}))$ for all $i \in [n]$. With such choice, we can see that $\exp(-\|x_i - x_j\|^2 / 2\tau_i^2)$ will be tightly concentrated around $\exp\left(-\frac{R^2_{\pi(i),\pi(j)}}{\min_{\ell \in [k]} R^2_{\pi(i),\ell}}\right)$,

a quantity that only depends on $\pi(i)$ and $\pi(j)$ - the clusters $x_i$ and $x_j$ belong to. Note that $e^{-100} \leq \exp\left(-\frac{R^2_{\pi(i),\pi(j)}}{\min_{\ell \in [k]} R^2_{\pi(i),\ell}}\right) \leq e^{-1}$, so $\exp(-\|x_i - x_j\|^2/2\tau_i^2)$'s are within a constant factor of each other. Then combined with the balancedness property, it is easy to see that the fact indeed holds. □

**Definition B.3** (Shrinkage parameter, gapped). *Let $\mathcal{X}$ be a balanced $O(d^{-\eta})$-regular dataset. Suppose we have $p_{ij} = \frac{c_{\pi(i),\pi(j)}}{n^2} \pm O\left(\frac{1}{n^2 d^\eta}\right)$ for all $i, j \in [n]$ ($i \neq j$) as in Fact B.2.*

*For each $\ell \in [k]$, define $\beta_\ell := \frac{1}{n} \sum_{\ell'=1}^{k} |\mathcal{C}_{\ell'}| \cdot c_{\ell,\ell'}$ to be the shrinkage parameter for cluster $\mathcal{C}_\ell$. We say that the shrinkage parameters $\beta_1, \ldots, \beta_k$ are gapped if the largest one among them is at least $1.1$ times larger than the second largest one, that is, letting $\ell^* = \mathrm{argmax}_{\ell \in [k]} \beta_\ell$, we have $\beta_{\ell^*} \geq 1.1 \max_{\ell \in [k] \setminus \{\ell^*\}} \beta_\ell$.*

**Remark.** *In the above definition, since $\beta_\ell$ is a convex combination of $\{c_{\ell,\ell'} : \ell' \in [k]\}$, we have $\beta_\ell = \Theta(1)$ for all $\ell \in [k]$.*

The following theorem says that if a dataset is balanced and $\gamma$-regular and has gapped shrinkage parameters, then the cluster with the largest shrinkage parameter will be approximately visible after running t-SNE for $\Theta(\log d)$ iterations.

**Theorem B.4.** *Let $\mathcal{X} = \{x_1, x_2, \ldots, x_n\} \subseteq \mathbb{R}^d$ be a balanced $\gamma$-regular dataset with $\mathcal{C}_1, \mathcal{C}_2, \ldots, \mathcal{C}_k$ defining the $k$ individual clusters, whose shrinkage parameters $\beta_1, \ldots, \beta_k$ are gapped. Suppose we have $n \geq k^{1+\omega} d^\omega$ for some constant $\omega > 0$, and let $\beta_1 = \max_{\ell \in [k]} \beta_\ell$ without loss of generality. Choose $h = 1$ and $\alpha = \rho n$ for a sufficiently small constant $\rho > 0$.*

*Let $\mathcal{Y}^{(T)}$ be the output of t-SNE (Algorithm 1) after $T = \Theta(\log d)$ iterations on input $\mathcal{X}$ with the above parameters. Then, with high probability over the choice of the initialization, $\mathcal{Y}^{(T)}$ is a $(1 - d^{-\Omega(1)})$-partial visualization of $\mathcal{X}$ where $\mathcal{C}_1$ is $(1 - d^{-\Omega(1)})$-visible.*

We prove Theorem B.4 in Appendix B.1. Then we prove Theorem 4.1 in Appendix B.2 by showing that the dataset $\mathcal{X}$ in Theorem 4.1 satisfies all the conditions in Theorem B.4.

## B.1 Proof of Theorem B.4

Recall that we have $p_{ij} = \frac{c_{\pi(i),\pi(j)}}{n^2} \pm O\left(\frac{1}{n^2 d^\eta}\right)$ for all $i, j \in [n]$ ($i \neq j$), and $\beta_\ell = \frac{1}{n} \sum_{\ell'=1}^{k} |\mathcal{C}_{\ell'}| \cdot c_{\ell,\ell'}$ for all $\ell \in [k]$ (Definition B.3). Let $C > 0$ be a constant such that $p_{ij} = \frac{c_{\pi(i),\pi(j)}}{n^2} \pm \frac{C}{n^2 d^\eta}$.

Let $\lambda_{ij}^{(t)} := h\left(\alpha p_{ij} - q_{ij}^{(t)}\right) q_{ij}^{(t)} Z^{(t)}$. Then the t-SNE update (3) can be written as

$$y_i^{(t+1)} = y_i^{(t)} + \sum_{j \neq i} \lambda_{ij}^{(t)} \left(y_j^{(t)} - y_i^{(t)}\right). \tag{18}$$

**Claim B.5.** *Under the setting of Theorem B.4, we have $y_i^{(t)} \in [-0.01, 0.01]^2$ for all $i \in [n]$ at any time $t$. As a consequence, at any time $t$ we have*

$$\left\|y_i^{(t)} - y_j^{(t)}\right\| \leq 0.03, \quad \forall i \neq j,$$

$$0.999 \leq q_{ij}^{(t)} Z^{(t)} \leq 1, \quad \forall i \neq j,$$

$$0.999 n(n-1) \leq Z^{(t)} \leq n(n-1),$$

$$\frac{0.999}{n(n-1)} \leq q_{ij}^{(t)} \leq \frac{1.002}{n(n-1)}, \quad \forall i \neq j,$$

$$\lambda_{ij}^{(t)} = (1 \pm 0.01) \frac{\alpha h c_{\pi(i),\pi(j)}}{n^2}, \quad \forall i \neq j. \tag{19}$$

*Proof.* We prove the claim by induction on $t$. First, at time $t = 0$, we have that the points $y_i^{(0)}$'s are initialized in a small region $[-0.01, 0.01]^2$.

Suppose we have $y_i^{(t)} \in [-0.01, 0.01]^2$ ($\forall i \in [n]$) at time $t$, similar to (13)-(15), we can get all the desired bounds in the claim except (19). To show (19), note that we have

$$\lambda_{ij}^{(t)} = h\left(\alpha\left(\frac{c_{\pi(i),\pi(j)}}{n^2} \pm \frac{C}{n^2 d^\eta}\right) - \frac{1 \pm 0.002}{n(n-1)}\right) \cdot (0.999 \pm 0.001), \qquad \forall i \neq j.$$

Since we have $\alpha = \Theta(n)$, it is easy to see that (19) holds for large $n$.

Furthermore, we have $\sum_{j \in [n] \setminus \{i\}} \lambda_{ij}^{(t)} \leq 1.01 \frac{\alpha h \max_{\ell \in [k]} \{c_{\pi(i),\ell}\}}{n}$; since we have $\alpha h = \rho n$ for a sufficiently small constant $\rho$, this sum is no more than 1 ($\forall i \in [n]$). Then from (18) we know that $y_i^{(t+1)}$ ($\forall i \in [n]$) is a convex combination of $\left\{y_j^{(t)} : j \in [n]\right\}$, so $y_i^{(t+1)}$ is in $[-0.01, 0.01]^2$ as well. By induction, we have completed the proof. $\square$

Our next step is to derive an upper bound on $\left|\lambda_{ij}^{(t)} - \lambda_{i'j}^{(t)}\right|$ for $i \sim i'$. Note that we trivially have $\left|\lambda_{ij}^{(t)} - \lambda_{i'j}^{(t)}\right| \leq 0.02 \frac{\alpha h c_{\pi(i),\pi(j)}}{n^2}$ according to (19), but this bound is not good enough for our purpose. Instead, we have the following lemma, which gives a stronger bound that depends on $\left\|y_i^{(t)} - y_{i'}^{(t)}\right\|$.

**Lemma B.6.** *Under the setting of Theorem B.4, for any $i, i', j \in [n]$ such that $i \sim i'$ and $j \notin \{i, i'\}$ and any $t$ we have*

$$\left|\lambda_{ij}^{(t)} - \lambda_{i'j}^{(t)}\right| \leq 2.002 C \frac{\alpha h}{n^2 d^\eta} + 0.1 \frac{\alpha h c_{\pi(i),\pi(j)}}{n^2} \left\|y_i^{(t)} - y_{i'}^{(t)}\right\|.$$

*Proof.* For ease of presentation we ignore the superscript "$(t)$" in the proof.

We have
$$\begin{aligned}\left|\lambda_{ij} - \lambda_{i'j}\right| &= \left|h(\alpha p_{ij} - q_{ij})q_{ij}Z - h(\alpha p_{i'j} - q_{i'j})q_{i'j}Z\right| \\ &\leq \alpha h \left|p_{ij} q_{ij} Z - p_{i'j} q_{i'j} Z\right| + h \left|q_{ij}^2 Z - q_{i'j}^2 Z\right|.\end{aligned} \qquad (20)$$

Let $d_{ij} := \|y_i - y_j\|$ and similarly $d_{i'j} := \|y_{i'} - y_j\|$. Using Claim B.5, we have

$$\begin{aligned}\left|q_{ij}Z - q_{i'j}Z\right| &= \left|\frac{1}{1+d_{ij}^2} - \frac{1}{1+d_{i'j}^2}\right| = \frac{\left|d_{ij}^2 - d_{i'j}^2\right|}{\left(1+d_{ij}^2\right)\left(1+d_{i'j}^2\right)} \\ &\leq \left|d_{ij}^2 - d_{i'j}^2\right| = (d_{ij} + d_{i'j})\left|d_{ij} - d_{i'j}\right| \\ &\leq (0.03 + 0.03)\left|d_{ij} - d_{i'j}\right| = 0.06 \left|\|y_i - y_j\| - \|y_{i'} - y_j\|\right| \\ &\leq 0.06 \|y_i - y_{i'}\|,\end{aligned}$$

Then we have
$$\begin{aligned}\left|q_{ij}^2 Z - q_{i'j}^2 Z\right| &= (q_{ij} + q_{i'j})\left|q_{ij}Z - q_{i'j}Z\right| \leq \left(\frac{1.002}{n(n-1)} + \frac{1.002}{n(n-1)}\right) \cdot 0.06 \|y_i - y_{i'}\| \\ &\leq \frac{0.13}{n(n-1)} \|y_i - y_{i'}\|\end{aligned} \qquad (21)$$

and

$$
\begin{aligned}
&\left| p_{ij}q_{ij}Z - p_{i'j}q_{i'j}Z \right| \\
&= \left| (p_{ij}q_{ij}Z - p_{i'j}q_{ij}Z) + (p_{i'j}q_{ij}Z - p_{i'j}q_{i'j}Z) \right| \\
&\leq \left| p_{ij}q_{ij}Z - p_{i'j}q_{ij}Z \right| + \left| p_{i'j}q_{ij}Z - p_{i'j}q_{i'j}Z \right| \\
&= q_{ij}Z \left| p_{ij} - p_{i'j} \right| + p_{i'j} \left| q_{ij}Z - q_{i'j}Z \right| \\
&\leq 1 \cdot \left| \left( \frac{c_{\pi(i),\pi(j)}}{n^2} \pm \frac{C}{n^2 d^\eta} \right) - \left( \frac{c_{\pi(i'),\pi(j)}}{n^2} \pm \frac{C}{n^2 d^\eta} \right) \right| + \left( \frac{c_{\pi(i'),\pi(j)}}{n^2} + \frac{C}{n^2 d^\eta} \right) \cdot 0.06 \left\| y_i - y_{i'} \right\| \quad (22) \\
&\leq \frac{2C}{n^2 d^\eta} + \left( \frac{c_{\pi(i),\pi(j)}}{n^2} + \frac{C}{n^2 d^\eta} \right) \cdot 0.06 \left\| y_i - y_{i'} \right\| \\
&\leq (2 + 0.06 \cdot 0.03) \frac{C}{n^2 d^\eta} + 0.06 \frac{c_{\pi(i),\pi(j)}}{n^2} \left\| y_i - y_{i'} \right\| \\
&\leq 2.002 \frac{C}{n^2 d^\eta} + 0.06 \frac{c_{\pi(i),\pi(j)}}{n^2} \left\| y_i - y_{i'} \right\|.
\end{aligned}
$$

Plugging (21) and (22) into (20), we get

$$
\begin{aligned}
\left| \lambda_{ij} - \lambda_{i'j} \right| &\leq \alpha h \left( 2.002 \frac{C}{n^2 d^\eta} + 0.06 \frac{c_{\pi(i),\pi(j)}}{n^2} \left\| y_i - y_{i'} \right\| \right) + h \frac{0.13}{n(n-1)} \left\| y_i - y_{i'} \right\| \\
&\leq 2.002 C \frac{\alpha h}{n^2 d^\eta} + 0.1 \frac{\alpha h c_{\pi(i),\pi(j)}}{n^2} \left\| y_i - y_{i'} \right\|,
\end{aligned}
$$

where in the last step we have used $\alpha = \Theta(n)$ which implies

$$
0.04 \frac{\alpha h c_{\pi(i),\pi(j)}}{n^2} \left\| y_i - y_{i'} \right\| \geq h \frac{0.13}{n(n-1)} \left\| y_i - y_{i'} \right\|
$$

for large $n$. $\square$

The following two lemmas say that for any $i, i' \in \mathcal{C}_\ell$, $y_i - y_{i'}$ approximately shrinks by a factor of $1 - \frac{\alpha h \beta_\ell}{n}$ after each iteration. This explains why we call $\beta_\ell$ the shrinkage parameter for cluster $\mathcal{C}_\ell$.

**Lemma B.7.** *Under the setting of Theorem B.4, for any $i, i' \in \mathcal{C}_\ell$ ($i \neq i'$) and any $t$, we have*

$$
y_i^{(t+1)} - y_{i'}^{(t+1)} = \left( 1 - (1 \pm 0.01) \frac{\alpha h \beta_\ell}{n} \right) \left( y_i^{(t)} - y_{i'}^{(t)} \right) + \xi_{ii'}^{(t)},
$$

*where $\left\| \xi_{ii'}^{(t)} \right\| \leq 0.07 C \frac{\alpha h}{n d^\eta} + 0.004 \frac{\alpha h \beta_\ell}{n} \left\| y_i^{(t)} - y_{i'}^{(t)} \right\|$.*

*Proof.* We have $y_i^{(t+1)} = y_i^{(t)} + \sum_{j \neq i} \lambda_{ij}^{(t)} \left( y_j^{(t)} - y_i^{(t)} \right)$ and $y_{i'}^{(t+1)} = y_{i'}^{(t)} + \sum_{j \neq i'} \lambda_{i'j}^{(t)} \left( y_j^{(t)} - y_{i'}^{(t)} \right)$. Subtracting the two we get

$$
\begin{aligned}
&y_i^{(t+1)} - y_{i'}^{(t+1)} \\
&= y_i^{(t)} - y_{i'}^{(t)} + \sum_{j \neq i} \lambda_{ij}^{(t)} \left( y_j^{(t)} - y_i^{(t)} \right) - \sum_{j \neq i'} \lambda_{i'j}^{(t)} \left( y_j^{(t)} - y_{i'}^{(t)} \right) \\
&= \left( 1 - 2\lambda_{ii'}^{(t)} \right) \left( y_i^{(t)} - y_{i'}^{(t)} \right) + \sum_{j \notin \{i,i'\}} \lambda_{ij}^{(t)} \left( y_j^{(t)} - y_i^{(t)} \right) - \sum_{j \notin \{i,i'\}} \lambda_{i'j}^{(t)} \left( y_j^{(t)} - y_{i'}^{(t)} \right) \\
&= \left( 1 - 2\lambda_{ii'}^{(t)} \right) \left( y_i^{(t)} - y_{i'}^{(t)} \right) + \sum_{j \notin \{i,i'\}} \left( \lambda_{ij}^{(t)} - \lambda_{i'j}^{(t)} \right) y_j^{(t)} - \left( \sum_{j \notin \{i,i'\}} \lambda_{ij}^{(t)} \right) y_i^{(t)} + \left( \sum_{j \notin \{i,i'\}} \lambda_{i'j}^{(t)} \right) y_{i'}^{(t)}
\end{aligned}
$$

$$= \left(1 - 2\lambda_{ii'}^{(t)} - \sum_{j \notin \{i,i'\}} \lambda_{ij}^{(t)}\right) \left(y_i^{(t)} - y_{i'}^{(t)}\right) + \sum_{j \notin \{i,i'\}} \left(\lambda_{ij}^{(t)} - \lambda_{i'j}^{(t)}\right) y_j^{(t)} + \left(\sum_{j \notin \{i,i'\}} \left(\lambda_{i'j}^{(t)} - \lambda_{ij}^{(t)}\right)\right) y_{i'}(t). \tag{23}$$

Note that from (19) we have $2\lambda_{ii'}^{(t)} + \sum_{j \notin \{i,i'\}} \lambda_{ij}^{(t)} = 2(1 \pm 0.01) \frac{\alpha h c_{\ell,\ell}}{n^2} + \sum_{j \notin \{i,i'\}} (1 \pm 0.01) \frac{\alpha h c_{\ell,\pi(j)}}{n^2} = (1 \pm 0.01) \sum_{j=1}^n \frac{\alpha h c_{\ell,\pi(j)}}{n^2} = (1 \pm 0.01) \frac{\alpha h \beta_\ell}{n}$. Hence from (23) we have

$$y_i^{(t+1)} - y_{i'}^{(t+1)} = \left(1 - (1 \pm 0.01)\frac{\alpha h \beta_\ell}{n}\right) \left(y_i^{(t)} - y_{i'}^{(t)}\right) + \xi_{ii'}^{(t)},$$

where $\xi_{ii'}^{(t)} = \sum_{j \notin \{i,i'\}} \left(\lambda_{ij}^{(t)} - \lambda_{i'j}^{(t)}\right) y_j^{(t)} + \left(\sum_{j \notin \{i,i'\}} \left(\lambda_{i'j}^{(t)} - \lambda_{ij}^{(t)}\right)\right) y_{i'}(t)$.

Finally, we can bound $\left\|\xi_{ii'}^{(t)}\right\|$ using Lemma B.6:

$$\begin{aligned}
\left\|\xi_{ii'}^{(t)}\right\| &\leq \sum_{j \notin \{i,i'\}} \left|\lambda_{ij}^{(t)} - \lambda_{i'j}^{(t)}\right| \cdot \left\|y_j^{(t)}\right\| + \sum_{j \notin \{i,i'\}} \left|\lambda_{ij}^{(t)} - \lambda_{i'j}^{(t)}\right| \cdot \left\|y_{i'}^{(t)}\right\| \\
&\leq \sum_{j \notin \{i,i'\}} \left|\lambda_{ij}^{(t)} - \lambda_{i'j}^{(t)}\right| \cdot 0.01\sqrt{2} + \sum_{j \notin \{i,i'\}} \left|\lambda_{ij}^{(t)} - \lambda_{i'j}^{(t)}\right| \cdot 0.01\sqrt{2} \\
&\leq 0.03 \sum_{j \notin \{i,i'\}} \left|\lambda_{ij}^{(t)} - \lambda_{i'j}^{(t)}\right| \\
&\leq 0.03 \sum_{j \notin \{i,i'\}} \left(2.002 C \frac{\alpha h}{n^2 d^\eta} + 0.1 \frac{\alpha h c_{\pi(i),\pi(j)}}{n^2} \left\|y_i^{(t)} - y_{i'}^{(t)}\right\|\right) \\
&\leq 0.07 C \frac{\alpha h}{n d^\eta} + 0.003 \sum_j \frac{\alpha h c_{\ell,\pi(j)}}{n^2} \left\|y_i^{(t)} - y_{i'}^{(t)}\right\| \\
&= 0.07 C \frac{\alpha h}{n d^\eta} + 0.003 \cdot (1 \pm 0.01) \frac{\alpha h \beta_\ell}{n} \left\|y_i^{(t)} - y_{i'}^{(t)}\right\| \\
&\leq 0.07 C \frac{\alpha h}{n d^\eta} + 0.004 \frac{\alpha h \beta_\ell}{n} \left\|y_i^{(t)} - y_{i'}^{(t)}\right\|.
\end{aligned}$$
□

**Lemma B.8.** *Under the setting of Theorem B.4, for any $i, i' \in \mathcal{C}_\ell$, if $\left\|y_i^{(t)} - y_{i'}^{(t)}\right\| \geq \frac{C'}{d^\eta}$ for some sufficiently large constant $C' > 0$, we have*

$$\left\|y_i^{(t+1)} - y_{i'}^{(t+1)}\right\| = \left(1 - (1 \pm 0.02)\frac{\alpha h \beta_\ell}{n}\right) \left\|y_i^{(t)} - y_{i'}^{(t)}\right\|;$$

*if $\left\|y_i^{(t)} - y_{i'}^{(t)}\right\| \leq \frac{C'}{d^\eta}$, we have $\left\|y_i^{(t+1)} - y_{i'}^{(t+1)}\right\| \leq \frac{C'}{d^\eta}$.*

*Proof.* From Lemma B.7 we have $y_i^{(t+1)} - y_{i'}^{(t+1)} = \left(1 - (1 \pm 0.01)\frac{\alpha h \beta_\ell}{n}\right) \left(y_i^{(t)} - y_{i'}^{(t)}\right) + \xi_{ii'}^{(t)}$, where $\left\|\xi_{ii'}^{(t)}\right\| \leq 0.07 C \frac{\alpha h}{n d^\eta} + 0.004 \frac{\alpha h \beta_\ell}{n} \left\|y_i^{(t)} - y_{i'}^{(t)}\right\|$.

If $\left\|y_i^{(t)} - y_{i'}^{(t)}\right\| \geq \frac{C'}{d^\eta}$, we have $\left\|\xi_{ii'}^{(t)}\right\| \leq 0.01 \frac{\alpha h \beta_\ell}{n} \left\|y_i^{(t)} - y_{i'}^{(t)}\right\|$, because we can ensure $\left\|y_i^{(t)} - y_{i'}^{(t)}\right\| \geq \frac{35}{3} \frac{C}{\beta_\ell d^\eta}$ when $C'$ is sufficiently large. Then we have

$$\left\|y_i^{(t+1)} - y_{i'}^{(t+1)}\right\| = \left(1 - (1 \pm 0.01)\frac{\alpha h \beta_\ell}{n}\right) \left\|y_i^{(t)} - y_{i'}^{(t)}\right\| \pm \left\|\xi_{ii'}^{(t)}\right\|$$

$$= \left(1 - (1 \pm 0.01)\frac{\alpha h \beta_\ell}{n}\right) \left\|y_i^{(t)} - y_{i'}^{(t)}\right\| \pm 0.01 \frac{\alpha h \beta_\ell}{n} \left\|y_i^{(t)} - y_{i'}^{(t)}\right\|$$

$$= \left(1 - (1 \pm 0.02)\frac{\alpha h \beta_\ell}{n}\right) \left\|y_i^{(t)} - y_{i'}^{(t)}\right\|.$$

If $\left\|y_i^{(t)} - y_{i'}^{(t)}\right\| \leq \frac{C'}{d^\eta}$, we have

$$\left\|y_i^{(t+1)} - y_{i'}^{(t+1)}\right\| = \left(1 - (1 \pm 0.01)\frac{\alpha h \beta_\ell}{n}\right) \left\|y_i^{(t)} - y_{i'}^{(t)}\right\| \pm \left\|\xi_{ii'}^{(t)}\right\|$$

$$\leq \left(1 - 0.99\frac{\alpha h \beta_\ell}{n}\right) \left\|y_i^{(t)} - y_{i'}^{(t)}\right\| + 0.07 C \frac{\alpha h}{n d^\eta} + 0.004 \frac{\alpha h \beta_\ell}{n} \left\|y_i^{(t)} - y_{i'}^{(t)}\right\|$$

$$\leq \left(1 - 0.9\frac{\alpha h \beta_\ell}{n}\right) \left\|y_i^{(t)} - y_{i'}^{(t)}\right\| + 0.07 C \frac{\alpha h}{n d^\eta}$$

$$\leq \left(1 - 0.9\frac{\alpha h \beta_\ell}{n}\right) \frac{C'}{d^\eta} + 0.07 C \frac{\alpha h}{n d^\eta}$$

$$\leq \frac{C'}{d^\eta},$$

where in the last step we use $0.9 \beta_\ell C' \geq 0.07 C$, which is true for sufficiently large $C'$. □

The following lemma is a key step in showing partial visualization of $\mathcal{C}_1$.

**Lemma B.9.** *Under the setting of Theorem B.4, there exist constants $\nu > 0$ and $C'' > 0$ such that after running t-SNE for $T = C'' \log d$ iterations, with high probability the followings are true:*

(a) $\left\|y_i^{(T)} - y_{i'}^{(T)}\right\| \leq \frac{C'}{d^\eta}$ for all $i, i' \in \mathcal{C}_1$;

(b) *for any $\ell \in [k] \setminus \{1\}$, there does not exist a subset $S_\ell \subseteq \mathcal{C}_\ell$ such that $|S_\ell| \geq d^{-\nu} |\mathcal{C}_\ell|$ and for all $i, i' \in S_\ell$, $\left\|y_i^{(T)} - y_{i'}^{(T)}\right\| \leq \frac{10 C'}{d^\eta}$.*

*Proof.* By our choices of $\alpha$ and $h$ we can ensure that $1 - (1 \pm 0.02)\frac{\alpha h \beta_\ell}{n}$ is in $(0, 1)$ for any $\ell \in [k]$; also note that $\frac{\alpha h \beta_\ell}{n} = \rho \beta_\ell$ is $\Theta(1)$.

Let $r_1 := 1 - 0.98 \frac{\alpha h \beta_1}{n} = 1 - 0.98 \rho \beta_1$ and $r_2 := 1 - 1.02 \frac{\alpha h \max_{\ell \in [k] \setminus \{1\}} \beta_\ell}{n} = 1 - 1.02 \rho \max_{\ell \in [k] \setminus \{1\}} \beta_\ell$. Then both $r_1$ and $r_2$ are constants in $(0, 1)$, and by the assumption $\beta_1 \geq 1.1 \max_{\ell \in [k] \setminus \{1\}} \beta_\ell$ we know that $r_1 < r_2$. From Lemma B.8 we know

$$\left\|y_i^{(t+1)} - y_{i'}^{(t+1)}\right\| \leq r_1 \left\|y_i^{(t)} - y_{i'}^{(t)}\right\|, \qquad \text{for } i, i' \in \mathcal{C}_1, \tag{24}$$

$$\left\|y_i^{(t+1)} - y_{i'}^{(t+1)}\right\| \geq r_2 \left\|y_i^{(t)} - y_{i'}^{(t)}\right\|, \qquad \text{for } i, i' \in \mathcal{C}_\ell, \ell \geq 2, \tag{25}$$

as long as $\left\|y_i^{(t)} - y_{i'}^{(t)}\right\| \geq \frac{C'}{d^\eta}$. This roughly means that cluster $\mathcal{C}_1$ is shrinking *strictly faster* than all other clusters.

Since $\left\|y_i^{(0)} - y_{i'}^{(0)}\right\| \leq 0.03$ for all $i, i' \in [n]$, from (24) we know that for all $i, i' \in \mathcal{C}_1$, we have $\left\|y_i^{(T)} - y_{i'}^{(T)}\right\| \leq \frac{C'}{d^\eta}$ after $T = \log_{1/r_1}\left(\frac{0.03}{\frac{C'}{d^\eta}}\right) = \Theta(\log d)$ iterations.[5] Hence after $T$ iterations every two points in cluster $\mathcal{C}_1$ have distance at most $\frac{C'}{d^\eta}$. Therefore (a) is proved.

---

[5] Note that according to Lemma B.8, once we have $\left\|y_i^{(t)} - y_{i'}^{(t)}\right\| \leq \frac{C'}{d^\eta}$ for some $t$, we will have $\left\|y_i^{(t')} - y_{i'}^{(t')}\right\| \leq \frac{C'}{d^\eta}$ for all $t' \geq t$.

Next we show (b). In particular, we show that for any $\ell \in [k] \setminus \{1\}$, with probability $1 - \exp\left(-n^{\Omega(1)}\right)$ there does not exist a subset $S_\ell \subseteq \mathcal{C}_\ell$ such that $|S_\ell| \geq d^{-\nu}|\mathcal{C}_\ell|$ and for all $i, i' \in S_\ell$, $\left\|y_i^{(T)} - y_{i'}^{(T)}\right\| \leq \frac{10C'}{d^\eta}$, where $\nu > 0$ is a sufficiently small constant. Then applying a union bound on $\ell \in [k] \setminus \{1\}$ we complete the proof of (b).

Suppose we have a subset $S_\ell \subseteq \mathcal{C}_\ell$ ($\ell \geq 2$) such that $\left\|y_i^{(T)} - y_{i'}^{(T)}\right\| \leq \frac{10C'}{d^\eta}$ for all $i, i' \in S_\ell$. Then from (25) we know that

$$
\begin{aligned}
\left\|y_i^{(0)} - y_{i'}^{(0)}\right\| &\leq \frac{10C'}{d^\eta}\left(\frac{1}{r_2}\right)^T = \frac{10C'}{d^\eta}\left(\frac{1}{r_2}\right)^{\log_{1/r_1}\left(\frac{0.03}{\frac{C'}{d^\eta}}\right)} \\
&= \frac{10C'}{d^\eta}\left(\frac{0.03}{\frac{C'}{d^\eta}}\right)^{\frac{\log\frac{1}{r_2}}{\log\frac{1}{r_1}}} = 0.3\left(\frac{0.03d^\eta}{C'}\right)^{\frac{\log\frac{1}{r_2}}{\log\frac{1}{r_1}}-1}.
\end{aligned}
\tag{26}
$$

Because $0 < r_1 < r_2 < 1$, the right hand side of (26) can be upper bounded by $d^{-\nu}$, where $\nu > 0$ is a constant. (Note that the exponent $\frac{\log\frac{1}{r_2}}{\log\frac{1}{r_1}} - 1$ is a negative constant.) We additionally make sure that $\nu \leq \frac{\omega}{1+\omega}$ (recall that $\omega$ is a constant specified in Theorem B.4 such that $n \geq k^{1+\omega}d^\omega$). Therefore we must have $\left\|y_i^{(0)} - y_{i'}^{(0)}\right\| \leq d^{-\nu}$ for all $i, i' \in S_\ell$. This means that all points in $\left\{y_i^{(0)} : i \in S_\ell\right\}$ are inside a small disk of radius $d^{-\nu}$. Since points $y_i^{(0)}$'s are generated i.i.d. from the uniform distribution over $[-0.01, 0.01]^2$, it is now straightforward to prove that $|S_\ell|$ cannot be too large. We have

$$
\begin{aligned}
&\Pr\left[\exists \text{ such subset } S_\ell \subseteq \mathcal{C}_\ell \text{ of size } \geq s\right] \\
&\leq \Pr\left[\exists s \text{ points in } \left\{y_i^{(0)} : i \in \mathcal{C}_\ell\right\} \text{ that are within distance } d^{-\nu} \text{ of each other}\right] \\
&\leq \binom{|\mathcal{C}_\ell|}{s}\left(\frac{\pi \cdot (d^{-\nu})^2}{0.02^2}\right)^{s-1} \leq \left(\frac{e|\mathcal{C}_\ell|}{s}\right)^s \left(\frac{8000}{d^{2\nu}}\right)^{s-1} = \frac{e|\mathcal{C}_\ell|}{s}\left(\frac{8000e|\mathcal{C}_\ell|}{sd^{2\nu}}\right)^{s-1}.
\end{aligned}
\tag{27}
$$

Let $s = \frac{|\mathcal{C}_\ell|}{d^\nu}$ in (27). From $|\mathcal{C}_\ell| = \Theta(n/k)$ we have $s = \Omega\left(\frac{n}{kd^\nu}\right) \geq \Omega\left(\frac{n}{kd^{\frac{\omega}{1+\omega}}}\right) = \Omega\left(\frac{n}{(k^{1+\omega}d^\omega)^{\frac{1}{1+\omega}}}\right) \geq \Omega\left(\frac{n}{n^{\frac{1}{1+\omega}}}\right) = n^{\Omega(1)}$ (recall $n \geq k^{1+\omega}d^\omega$ and $\nu \leq \frac{\omega}{1+\omega}$). Then (27) implies

$$
\Pr\left[\exists \text{ such subset } S_\ell \subseteq \mathcal{C}_\ell \text{ of size } \geq \frac{|\mathcal{C}_\ell|}{d^\nu}\right] \leq ed^\nu \left(\frac{8000e}{d^\nu}\right)^{n^{\Omega(1)}} \leq \exp\left(-n^{\Omega(1)}\right).
$$

Therefore the proof is completed. $\square$

Now we can finish the proof of Theorem B.4 using Lemma B.9.

*Proof of Theorem B.4.* Suppose that the high-probability events in Lemma B.9 happen. From Lemma B.9 we know that $\left\|y_i^{(T)} - y_{i'}^{(T)}\right\| \leq C'd^{-\eta}$ for all $i, i' \in \mathcal{C}_1$. Consider a fixed $i_1 \in \mathcal{C}_1$.

According to Lemma B.9, we have $\left|\left\{j \in \mathcal{C}_\ell : y_j^{(T)} \in \mathcal{B}\left(y_{i_1}^{(T)}, 5C'd^{-\eta}\right)\right\}\right| \leq d^{-\nu}|\mathcal{C}_\ell|$ for all $\ell \in [k] \setminus \{1\}$, because for $y_j^{(T)}, y_{j'}^{(T)} \in \mathcal{B}\left(y_{i_1}^{(T)}, 5C'd^{-\eta}\right)$ we must have $\left\|y_j^{(T)} - y_{j'}^{(T)}\right\| \leq 10C'd^{-\eta}$.

Let $\mathcal{P}_{\text{err}} = \bigcup_{\ell=2}^{k} \left\{ j \in \mathcal{C}_\ell : y_j^{(T)} \in \mathcal{B}\left(y_{i_1}^{(T)}, 5C'd^{-\eta}\right) \right\}$. Then we have $|\mathcal{P}_{\text{err}}| \leq \sum_{\ell=2}^{k} d^{-\nu}|\mathcal{C}_\ell| < d^{-\nu}n$. By the definition of $\mathcal{P}_{\text{err}}$ we know that for any $j \in [n] \setminus (\mathcal{C}_1 \cup \mathcal{P}_{\text{err}})$ we have $\left\| y_j^{(T)} - y_{i_1}^{(T)} \right\| > 5C'd^{-\eta}$. Therefore, for any $i, i' \in \mathcal{C}_1$ and any $j \in [n] \setminus (\mathcal{C}_1 \cup \mathcal{P}_{\text{err}})$ we have $\left\| y_i^{(T)} - y_{i'}^{(T)} \right\| \leq C'd^{-\eta}$ and $\left\| y_i^{(T)} - y_j^{(T)} \right\| \geq \left\| y_j^{(T)} - y_{i_1}^{(T)} \right\| - \left\| y_i^{(T)} - y_{i_1}^{(T)} \right\| > 5C'd^{-\eta} - C'd^{-\eta} = 4C'd^{-\eta}$. This means that $\mathcal{C}_1$ is $d^{-\nu}$-approximated visualized by itself in $\mathcal{Y}^{(T)}$. The proof is completed. $\square$

## B.2 Proof of Theorem 4.1

*Proof of Theorem 4.1.* It suffices to check that all the conditions in Theorem B.4 are satisfied. We have $n > 4d$, so $n \geq k^{1+\omega}d^\omega$ is satisfied with $\omega = 1$. Since $|\mathcal{C}_1| = |\mathcal{C}_2| = n/2$, balancedness is satisfied. Standard distance concentration for Gaussians tell us that, with high probability, $\|x_i - x_j\|^2 = \left(1 \pm O(\sqrt{(\log n)/d})\right) R^2_{\pi(i),\pi(j)}$ for all $i \neq j$, where

$$R^2_{1,1} = 2\sigma_1^2$$
$$R^2_{2,2} = 2\sigma_2^2$$
$$R^2_{1,2} = R^2_{2,1} = \sigma_1^2 + \sigma_2^2.$$

Thus $\mathcal{X}$ is $O(d^{-\eta})$-regular for any constant $\eta < 1/2$ with high probability. Also note that since $1.5 \leq \sigma_2/\sigma_1 \leq 10$ we have $R_{1,1} < R_{1,2} < R_{2,2} \leq 10R_{1,1}$.

It remains to show that the shrinkage parameters satisfy $\beta_1 \geq 1.1\beta_2$. For this we need to compute $\beta_1$ and $\beta_2$.

For $i \in \mathcal{C}_1$ we have $2\tau_i^2 = \min_{j \neq i} \|x_i - x_j\|^2 = (1 \pm O(d^{-\eta})) \cdot R^2_{1,1}$. Thus for $i, j \in \mathcal{C}_1$ ($i \neq j$) we have

$$\exp\left(-\frac{\|x_i - x_j\|^2}{2\tau_i^2}\right) = \exp\left(-\frac{(1 \pm O(d^{-\eta})) \cdot R^2_{1,1}}{(1 \pm O(d^{-\eta})) \cdot R^2_{1,1}}\right) = \exp\left(-1 \pm O(d^{-\eta})\right) = e^{-1}(1 \pm O(d^{-\eta})),$$

and similarly for $i \in \mathcal{C}_1, j' \in \mathcal{C}_2$ we have

$$\exp\left(-\frac{\|x_i - x_{j'}\|^2}{2\tau_i^2}\right) = \exp\left(-\frac{(1 \pm O(d^{-\eta})) \cdot R^2_{1,2}}{(1 \pm O(d^{-\eta})) \cdot R^2_{1,1}}\right) = \exp\left(-\frac{R^2_{1,2}}{R^2_{1,1}} \pm O(d^{-\eta})\right)$$
$$= e^{-R^2_{1,2}/R^2_{1,1}}(1 \pm O(d^{-\eta})).$$

Let $Z_i = \sum_{j \neq i} \exp\left(-\frac{\|x_i - x_{j'}\|^2}{2\tau_i^2}\right)$. We know that for $i \in \mathcal{C}_1$ we have

$$Z_i = \left(\frac{n}{2} - 1\right) e^{-1}(1 \pm O(d^{-\eta})) + \frac{n}{2} e^{-R^2_{1,2}/R^2_{1,1}}(1 \pm O(d^{-\eta}))$$
$$= \frac{n}{2}(1 \pm O(d^{-\eta})) \left(e^{-1} + e^{-R^2_{1,2}/R^2_{1,1}}\right).$$

Therefore we have

$$p_{j|i} = \frac{\exp\left(-\frac{\|x_i - x_{j'}\|^2}{2\tau_i^2}\right)}{Z_i} = \begin{cases} \frac{2}{n} \cdot \frac{e^{-1}}{e^{-1} + e^{-R^2_{1,2}/R^2_{1,1}}} \cdot (1 \pm O(d^{-\eta})), & i, j \in \mathcal{C}_1, i \neq j, \\ \frac{2}{n} \cdot \frac{e^{-R^2_{1,2}/R^2_{1,1}}}{e^{-1} + e^{-R^2_{1,2}/R^2_{1,1}}} \cdot (1 \pm O(d^{-\eta})), & i \in \mathcal{C}_1, j \in \mathcal{C}_2. \end{cases}$$

Similarly, we repeat the calculation for $\mathcal{C}_2$ and get

$$p_{j|i} = \begin{cases} \frac{2}{n} \cdot \frac{e^{-1}}{e^{-1}+e^{-R_{2,2}^2/R_{1,2}^2}} \cdot (1 \pm O(d^{-\eta})), & i \in \mathcal{C}_2, j \in \mathcal{C}_1, \\ \frac{2}{n} \cdot \frac{e^{-R_{2,2}^2/R_{1,2}^2}}{e^{-1}+e^{-R_{2,2}^2/R_{1,2}^2}} \cdot (1 \pm O(d^{-\eta})), & i,j \in \mathcal{C}_2, i \neq j. \end{cases}$$

Then form $p_{ij} = \frac{p_{i|j}+p_{j|i}}{2n}$ we have

$$p_{ij} = \begin{cases} \frac{1}{n^2} \cdot \frac{2e^{-1}}{e^{-1}+e^{-R_{1,2}^2/R_{1,1}^2}} \cdot (1 \pm O(d^{-\eta})), & i,j \in \mathcal{C}_1, i \neq j, \\ \frac{1}{n^2} \cdot \left( \frac{e^{-R_{1,2}^2/R_{1,1}^2}}{e^{-1}+e^{-R_{1,2}^2/R_{1,1}^2}} + \frac{e^{-1}}{e^{-1}+e^{-R_{2,2}^2/R_{1,2}^2}} \right) \cdot (1 \pm O(d^{-\eta})), & i \in \mathcal{C}_1, j \in \mathcal{C}_2, \\ \frac{1}{n^2} \cdot \frac{2e^{-R_{2,2}^2/R_{1,2}^2}}{e^{-1}+e^{-R_{2,2}^2/R_{1,2}^2}} \cdot (1 \pm O(d^{-\eta})), & i,j \in \mathcal{C}_2, i \neq j. \end{cases}$$

This means

$$c_{1,1} = \frac{2e^{-1}}{e^{-1}+e^{-R_{1,2}^2/R_{1,1}^2}},$$

$$c_{1,2} = c_{2,1} = \frac{e^{-R_{1,2}^2/R_{1,1}^2}}{e^{-1}+e^{-R_{1,2}^2/R_{1,1}^2}} + \frac{e^{-1}}{e^{-1}+e^{-R_{2,2}^2/R_{1,2}^2}},$$

$$c_{2,2} = \frac{2e^{-R_{2,2}^2/R_{1,2}^2}}{e^{-1}+e^{-R_{2,2}^2/R_{1,2}^2}},$$

for $p_{ij} = \frac{c_{\pi(i),\pi(j)}}{n^2} \pm O\left(\frac{1}{n^2 d^\eta}\right)$ to hold.

Now we have $\beta_1 = \frac{1}{n}(|\mathcal{C}_1|c_{1,1} + |\mathcal{C}_2|c_{1,2}) = \frac{1}{2}(c_{1,1} + c_{1,2})$ and $\beta_2 = \frac{1}{2}(c_{1,2} + c_{2,2})$. Observe that $\beta_1 + \beta_2 = 2$. Thus to show $\beta_1 \geq 1.1\beta_2$ it suffices to show $\beta_1 \geq 1.05$. We have

$$\beta_1 = \frac{1}{2}\left( \frac{2e^{-1}}{e^{-1}+e^{-R_{1,2}^2/R_{1,1}^2}} + \frac{e^{-R_{1,2}^2/R_{1,1}^2}}{e^{-1}+e^{-R_{1,2}^2/R_{1,1}^2}} + \frac{e^{-1}}{e^{-1}+e^{-R_{2,2}^2/R_{1,2}^2}} \right)$$

$$= \frac{1}{2}\left( 1 + \frac{e^{-1}}{e^{-1}+e^{-R_{1,2}^2/R_{1,1}^2}} + \frac{e^{-1}}{e^{-1}+e^{-R_{2,2}^2/R_{1,2}^2}} \right).$$

Note that $\frac{R_{1,2}^2}{R_{1,1}^2} \cdot \frac{R_{2,2}^2}{R_{1,2}^2} = \frac{R_{2,2}^2}{R_{1,1}^2} = \frac{\sigma_2^2}{\sigma_1^2} \geq 1.5^2$. So we must have $\max\left\{\frac{R_{1,2}^2}{R_{1,1}^2}, \frac{R_{2,2}^2}{R_{1,2}^2}\right\} \geq 1.5$. We also have $\min\left\{\frac{R_{1,2}^2}{R_{1,1}^2}, \frac{R_{2,2}^2}{R_{1,2}^2}\right\} \geq 1$. Then we can bound $\beta_1$ as

$$\beta_1 \geq \frac{1}{2}\left(1 + \frac{e^{-1}}{e^{-1}+e^{-1}} + \frac{e^{-1}}{e^{-1}+e^{-1.5}}\right) > 1.05.$$

This completes the proof. □

32